%% file: main.tex
\documentclass[runningheads]{llncs}

\usepackage[mobile]{eccv}

\usepackage{eccvabbrv}

\usepackage{graphicx}
\usepackage{booktabs}

\usepackage[accsupp]{axessibility}  

\usepackage{hyperref}

\usepackage{orcidlink}

\input{custom_cmds}

\usepackage{pdfpages}

\begin{document}

\title{Enhancing Diffusion Models with Text-Encoder Reinforcement Learning}

\titlerunning{TexForce}

\author{Chaofeng Chen\inst{1}\thanks{These authors contributed equally to this work.}\and
Annan Wang\inst{1}$^\star$\and
Haoning Wu\inst{1} \and
Liang Liao\inst{1} \and \\
Wenxiu Sun\inst{3} \and
Qiong Yan\inst{3}\and
Weisi Lin\inst{2}\,\Envelope 
}

\authorrunning{C.~Chen, A.~Wang et al.}

\institute{S-Lab, Nanyang Technological University \and
CCDS, Nanyang Technological University \ \ \ \ 
\inst{3} Sensetime Research \\
\email{chaofeng.chen@ntu.edu.sg} \ \email{c190190@e.ntu.edu.sg} \ \email{wslin@ntu.edu.sg}\\
\url{https://github.com/chaofengc/TexForce}
}

\maketitle
\begin{center}
    \centering
    \newcommand{\imgwidth}{.33\textwidth}
    \includegraphics[width=\linewidth]{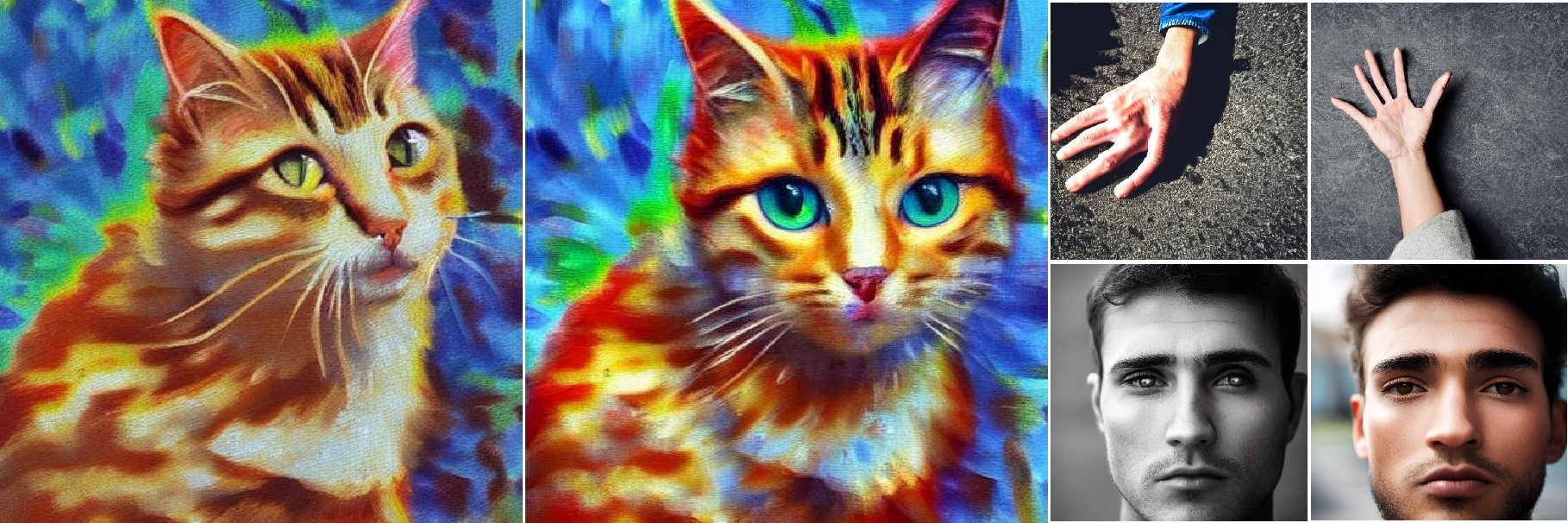}
    \makebox[0.31\textwidth]{(a) SDv1.4}
    \makebox[0.31\textwidth]{(b) TexForce (SDv1.4)}
    \makebox[0.16\linewidth]{(c) SDv1.4}
    \makebox[0.16\linewidth]{(d) TexForce}
    
    \captionof{figure}{By refining the text encoder through reinforcement learning, the proposed \textbf{\emph{TexForce}} with Stable Diffusion v1.4 can generate images that align better with human quality preference. \emph{The compared images are generated with the same seed and prompts. (a)(b): ``Impressionist painting of a cat, high quality''; (c)(d): ``A photo of a hand'' \& ``A complete face of a man''.}} \label{fig:teaser}
\end{center}%

\begin{abstract}
  Text-to-image diffusion models are typically trained to optimize the log-likelihood objective, which presents challenges in meeting specific requirements for downstream tasks, such as image aesthetics and image-text alignment. Recent research addresses this issue by refining the diffusion U-Net using human rewards through reinforcement learning or direct backpropagation. However, many of them overlook the importance of the text encoder, which is typically pretrained and fixed during training. In this paper, we demonstrate that by finetuning the text encoder through reinforcement learning, we can enhance the text-image alignment of the results, thereby improving the visual quality. Our primary motivation comes from the observation that the current text encoder is suboptimal, often requiring careful prompt adjustment. While fine-tuning the U-Net can partially improve performance, it remains suffering from the suboptimal text encoder. Therefore, we propose to use reinforcement learning with low-rank adaptation to finetune the text encoder based on task-specific rewards, referred as \textbf{TexForce}. We first show that finetuning the text encoder can improve the performance of diffusion models. Then, we illustrate that TexForce can be simply combined with existing U-Net finetuned models to get much better results without additional training. Finally, we showcase the adaptability of our method in diverse applications, including the generation of high-quality face and hand images.
\end{abstract}

Generative models have witnessed notable advancements in recent years, transitioning from earlier Generative Adversarial Networks (GAN) \cite{goodfellow2020generative, brock2018largebiggan, karras2019style} to the more recent diffusion models \cite{ho2020denoisingddpm, dhariwal2021diffusion, ddim}. Text-to-image models like Stable Diffusion \cite{stablediffusion}, DALLE \cite{dalle1, dalle2}, and Imagen \cite{imagen}, trained on extensive datasets, have demonstrated impressive capabilities in producing high-quality images from textual prompts. However, these diffusion models primarily optimize the log-likelihood objective, which, although effective for generative tasks, may not consistently fulfill specific requirements for downstream applications. Key challenges include achieving desirable image aesthetics and aligning generated images with text descriptions, both of which are critical for applications in areas such as content generation and multimedia synthesis.

\begin{wrapfigure}{r}{0.4\textwidth}
    \centering
    \includegraphics[width=1.0\linewidth]{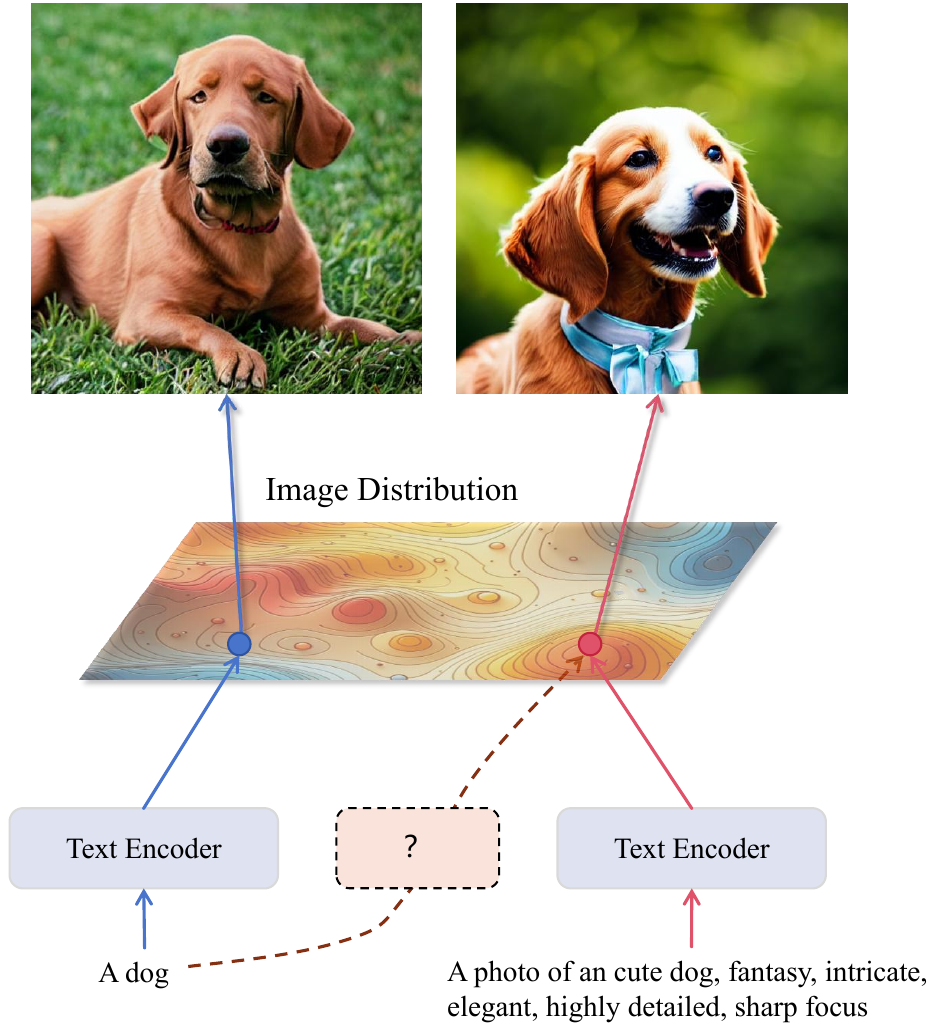}
    \caption{The qualities of outputs from pretrained diffusion models vary a lot with different prompts. Through the reinforcement learning, we can finetune text encoder to better align with images.}
    \label{fig:intro_motivation}
\end{wrapfigure}

Although prompt engineering is helpful in some cases, such as \cref{fig:intro_motivation}, these techniques \cite{witteveen2022investigating,wang2022diffusiondb,hao2022optimizing} have inherent limitations, including a lack of precise control, limited generalization across various models, and inadequacy in addressing complex demands. For instance, current models encounter difficulties in generating visual texts \cite{liu2022character}, and comprehending object counts \cite{tifa,lee2023aligning}. Therefore, recent efforts draw inspiration from the success of \emph{reinforcement learning from human feedback (RLHF)} employed in large language models \cite{ouyang2022training} and adopt similar strategies to enhance the alignment capabilities of diffusion models \cite{lee2023aligning,2023DPOK,black2023ddpo,prabhudesai2023alignprop}. While these methods show promise, they all finetune the U-Net conditioned on the fixed suboptimal text encoder, which constrains their efficacy. 

In this paper, we introduce \textbf{TexForce}, an innovative method that applies reinforcement learning combined with low-rank adaptation to enhance the text encoder using task-specific rewards. We utilize the DDPO (denoising diffusion policy optimization) \cite{black2023ddpo} algorithm to update the text encoder, which is based on PPO (proximal policy optimization \cite{ppo}) in the iterative denoising process. Unlike direct backpropagation, this RL algorithm does not require differentiable rewards, offering greater flexibility. By finetuning with LoRA \cite{lora}, TexForce can adapt to diverse tasks by simply switching the LoRA weights, and also allows for the fusion of different LoRA weights to combine the capabilities learned from different rewards. Most importantly, TexForce can be seamlessly integrated with existing finetuned U-Net models from previous methods and achieves much better performance without additional training. 

As illustrated in \cref{fig:teaser}, our approach significantly enhances the result quality of Stable Diffusion 1.4 (SDv1.4) \cite{stablediffusion} when finetuned to align with different rewards. We validate our approach through extensive experiments on both single-prompt and multi-prompt tasks across various reward functions. Moreover, we showcase the adaptability of our method across various applications, including the generation of high-quality face and hand images. Our contributions can be summarized as follows:
\begin{itemize}
    \item We observe that when optimizing reinforcement learning (RL) rewards, finetuning the U-Net component of diffusion models carries the risk of compromising image appearance, whereas finetuning the text encoder mitigates such concerns and preserves semantics better.
    \item We find that it is possible to directly combine the LoRA weights from text encoder and U-Net \emph{without extra training}. This straightforward fusion addresses the challenges associated with finetuned U-Net while upholding the merits of finetuned text encoders.
\end{itemize}

\section{Related Works}

\subsection{Text-to-Image Diffusion Models}

Denoising diffusion models \cite{sohl2015deep,ho2020denoisingddpm,song2020improved,song2021maximum} have become the de facto standard for generative tasks, owing to their remarkable capabilities in generating diverse multimedia content, including images \cite{dhariwal2021diffusion,stablediffusion}, videos \cite{guo2023animatediff,text2video-zero,wu2023tune}, 3D content \cite{poole2022dreamfusion,liu2023zero}, and more. Text-to-image models, particularly those creating images based on textual prompts, have gained significant traction attributable to the availability of powerful models such as StableDiffusion \cite{stablediffusion}, DALLE \cite{dalle1,dalle2} and Imagen \cite{imagen}. Several approaches have emerged to enhance control over texture details in the generated outputs. Noteworthy methods like DreamBooth \cite{ruiz2023dreambooth} and Texture Inversion \cite{gal2022textureinversion} offer tailored solutions for specific image requirements. To improve the generalization capabilities, ControlNet \cite{zhang2023controlnet}, T2I-Adapter \cite{mou2023t2i} and $\mathcal{W}_+$-Adapter \cite{li2023w-plus-adapter} introduce additional image encoders to control the structure and details. Nonetheless, they still require a large number of paired images to train, and may struggle to meet the diverse demands of various tasks. Prompt engineering \cite{hao2022optimizing} is another popular approach aimed at enhancing the quality of generated images. However, this method is constrained by the expressiveness of text prompts and pretrained models and may not be straightforward when addressing complex tasks such as aesthetic quality and object composition \cite{lee2023aligning,2023DPOK}. 

\subsection{Learning from Feedback in Diffusion Models}

Recent efforts have aimed to optimize diffusion models using human rewards or task objectives, typically categorized into three main approaches: reward-weighted regression (RWR), reinforcement learning (RL), and direct backpropagation. RWR methods like RAFT \cite{dong2023raft}, Lee et al. \cite{lee2023aligning}, and Emu \cite{dai2023emu} start by assessing image quality with human feedback and then re-weight or select high-quality examples to enhance performance. RL methods, exemplified by DDPO \cite{black2023ddpo} and DPOK \cite{2023DPOK}, treat the denoising process as a Markov decision process and optimize the model using RL algorithms, such as PPO \cite{ppo}. Direct backpropagation methods, including AlignProp \cite{prabhudesai2023alignprop}, ReFL \cite{imagereward}, and DRaFT \cite{clark2023draft}, propagate gradients directly from the reward function to the model. Because these models only finetune U-Net conditioned on the suboptimal text encoder, their effectiveness in aligning outputs with text prompts is often limited. A concurrent work, TextCraftor \cite{li2024textcraftor}, also explores fine-tuning the text encoder; however, it relies on direct backpropagation and is incompatible with non-differentiable rewards.

\subsection{Quality Metrics for Generative Models}

With the increasing popularity of generative models, several benchmarks \cite{hps, imagereward, pickscore, agiqa3k, wu2023qbench} have been developed to evaluate the quality of generated images. Notably, ImageReward \cite{imagereward}, PickScore \cite{pickscore}, and HPS \cite{hps} are among the more frequently employed benchmarks. Additionally, image aesthetic metrics, particularly the LAION Aesthetics Predictor \cite{laionaes}, find widespread application in data filtering \cite{stablediffusion} and results assessment.

\section{Method}

\subsection{Preliminaries on Diffusion Models} \label{sec:preliminaries}

Diffusion models \cite{ho2020denoisingddpm, stablediffusion} belong to the class of generative models that leverage noise-driven processes to progressively transform data distributions. This process contains a controlled noise addition phase (forward diffusion) and a noise removal phase (reverse diffusion). Given image samples $\xo$ originating from the data distribution $q(\xo)$, the forward diffusion process generates a sequence of images $\{\mx_t\}_{t=1}^T$ by iteratively introducing noise via a Markov chain with a predefined noise schedule. Then, the reverse diffusion process is to learn a denoising U-Net $\epsilon_\theta$ to estimate the cleaner $\mx_{t-1}$ with the noisy $\mx_t$:
\begin{equation}
    p_\theta(\mx_{t-1}|\mx_t) = \nd(\mx_{t-1}; \mathbf{\mu}_\theta(\mx_t, t), \mathbf{\Sigma}_\theta(\mx_t, t)),
\end{equation}
where $\theta$ is the learnable parameter. For text-to-image diffusion models \cite{stablediffusion}, this process is conditioned on a text input $s$, encoded with a text encoder $\mz = \tau_\phi(s)$. Then the network $\epsilon_\theta$ is trained with the following objective:
\begin{equation}
    \mathcal{L}(\theta) = \mathbb{E}_{\mx_t, s, t, \epsilon \sim \nd(0, \mathbf{I})} \left[ \| \epsilon - \epsilon_\theta(\mx_t, t, \mz) \|_2^2 \right], \label{eq:denoising_loss}
\end{equation}
which aims to optimize the variational lower bound on the log-likelihood of the data distribution $q(\xo)$. It is worth noting that the text encoder $\tau_\phi$ is usually a pretrained model, such as CLIP \cite{clip}, and is fixed during training.

\subsection{Reinforcement Learning with LoRA}

According to the above formulation, diffusion models are learnt to optimize the log-likelihood objective \cref{eq:denoising_loss}, which is not directly related to the task requirements. 
To address this issue, we propose to finetune the text encoder $\tau_\phi$ with reinforcement learning (RL). Details come as follows.

\para{RL in Diffusion Models.} 
In our setting, the RL framework optimizes the policy defined by the diffusion model conditioned on the text embeddings. 
\begin{wrapfigure}{r}{0.5\textwidth}
    \centering
    \includegraphics[width=1.0\linewidth]{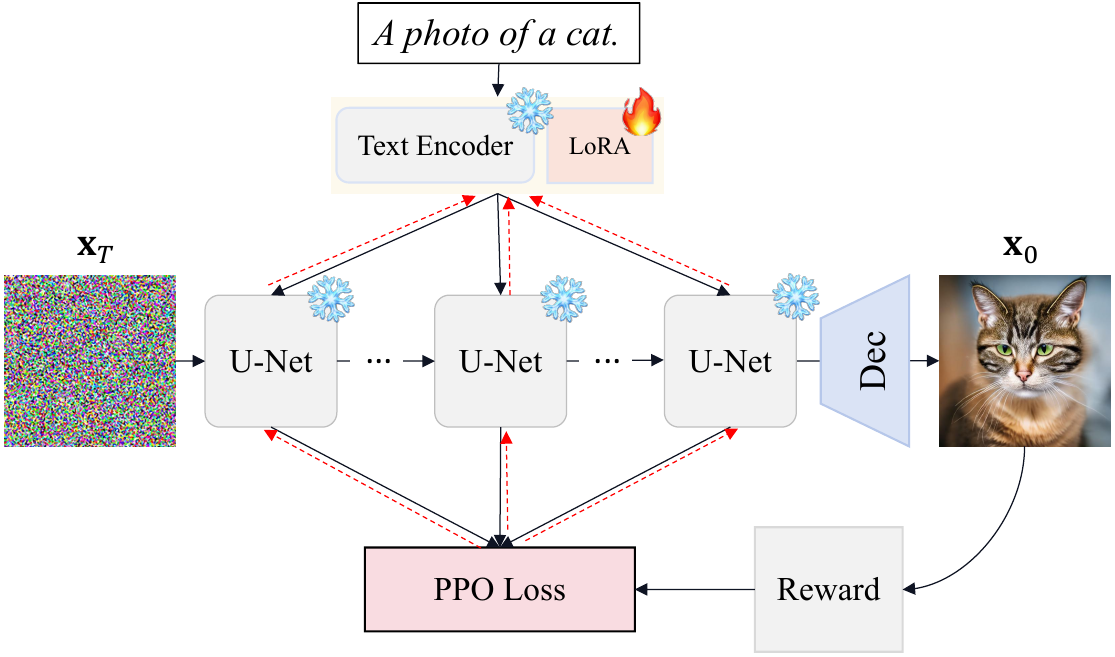}
    \caption{Illustration of text encoder finetune with PPO algorithm.}
\end{wrapfigure}
The text encoder $\tau_\phi$ acts as the policy network that maps text descriptions to actions (text embeddings), which then influences the generative process of the diffusion model. Let $R$ be the reward function that evaluates the quality of the generated images, which could encapsulate various aspects, such as image-text alignment and image quality, and adherence to specific attributes desired in the output. Then the objective of RL is to maximize the expected reward:
\begin{equation}
    J(\phi) = \mathbb{E} \left[R(\xo, s) \right]. \label{eq:rl_objective}
\end{equation}
Since the denoising process can be formulated as a Markov decision process \cite{black2023ddpo}, \ie, $p_\theta(\xo|\mz) = p(\mx_T)\prod_{t=1}^T p_\theta(\mx_{t-1}|\mx_t, \mz)$, the policy gradient of \cref{eq:rl_objective} can be computed as:
\begin{equation}
    \nabla_\phi J = \mathbb{E}\left[ \sum_{t=0}^T \nabla_\phi \log p_\theta(\mx_{t-1}|\mx_t, \tau_\phi(s)) R(\xo, s) \right]. \label{eq:rl_gradient}
\end{equation}
Following the DDPO algorithm, we use the Proximal Policy Optimization (PPO) \cite{ppo} to keep stable learning dynamics. It applies importance sampling with clipped probability ratio to \cref{eq:rl_objective} which becomes:
\begin{equation}
    J = \mathbb{E}\left[ \min(r_t(\phi)A, \text{clip}(r_t(\phi), 1 - \lambda, 1 + \lambda)A) \right], \label{eq:rl_ppo} 
\end{equation}
where the advantage value $A$ is the normalized rewards $R$ over a buffer set of $\xo$, and $r_t$ is the probability ratio between the new policy and the old policy for the denoise step $p_\theta(\mx_{t-1}|\mx_t, \tau_\phi(s))$. Since the policy is an isotropic Gaussian, the probability can be easily calculated. Then, we can calculate the gradient for \cref{eq:rl_ppo} similar to \cref{eq:rl_gradient} to update the policy network $\tau_\phi$. More details are provided in supplementary materials.

\para{Low-Rank Adaptation (LoRA)} \cite{lora} is a technique that allows for the modification of large pre-trained models without the need for extensive re-training. It achieves this by inserting trainable low-rank matrices into the original feed-forward layer as $W' = W + \alpha \Delta W$, where $\Delta W$ is the learnable weights initialized to zero and $\alpha$ is a scale factor. Such low-rank weight matrices are shown to be helpful in preventing the model from overfitting to the training data \cite{lora}.  

\subsection{Discussion of Finetuning for Diffusion Model}

In this part, we briefly discuss the advantages of finetuning the text encoder with reinforcement learning to improve the performance of diffusion models.

\para{Finetune of Diffusion Model.} As discussed in \cref{sec:preliminaries}, given the text $s$ and $\xo$, the denoising network $\epsilon_\theta$ is learned by maximizing the following lower bound:
\begin{equation}
    \mathbb{E}_{\mz\sim q_\phi(\mz|s)} \left[ \log(p_\theta(\mx_{0:T}|\mz))\right] - D_{KL}(q_\phi(\mz|s)||p(\mz)). \label{eq:elbo}
\end{equation}
In the training stage of diffusion models, $\phi$ is usually fixed and $p_\theta(\xo|\mz)$ are learned through classifier free guidance \cite{ho2022classifier}. With an extremely large amount of $s$ in datasets such as LAION-400M \cite{laion-5b,laion400m}, it is reasonable to assume that $q_\phi$ is a good estimation of $p(\mz)$ even when $\phi$ is fixed. However, in the finetuning stage, we expect to use a small amount of $s$ to optimize \cref{eq:elbo} for specific tasks. In such cases, the $q_\phi$ is likely to be a suboptimal estimation of $p(\mz)$, and thus largely increasing the second KL term. Therefore, we believe that it is necessary to finetune the text encoder $\tau_\phi$ to minimize the second term when the finetune dataset is limited.

\para{RL v.s. Direct Backpropagation.} Besides reinforcement learning, recent approaches also directly backpropagate the gradients through the denoising steps \cite{imagereward,prabhudesai2023alignprop,clark2023draft}: $\nabla_\theta \mathcal{L} = \sum_{m}^n \frac{\partial R}{\partial\mx_t} \frac{\partial \mx_t}{\partial \theta}$, where $m \leq n \in [0, T]$. However, this approach is more likely to overfit the reward function and lead to mode collapse. For instance, in DRaFT \cite{clark2023draft}, the model may collapse to generate a single image to achieve high aesthetic rewards. Besides, RL does not require differentiable quality rewards, and is much more flexible than direct backpropagation. For example, current applications can collect human feedbacks and use them as rewards to directly finetune the model.

\section{Experiments}

\subsection{Implementation Details}

\para{Prompt Datasets.} We follow previous works \cite{black2023ddpo,prabhudesai2023alignprop,2023DPOK,imagereward,hps} and use three types of prompt datasets with their corresponding experimental settings:
\begin{itemize}
    \item \textbf{Simple animal prompts \cite{black2023ddpo}.} A simple dataset with a curated list of 45 common animals for training. 
    \item \textbf{Single phrases.} Four single phrases from DPOK \cite{2023DPOK} to test the model capabilities under different scenarios.
    \item \textbf{Complex long prompts.} Subsets from ImageReward \cite{imagereward} and HPSv2 \cite{hps}. The former contains $20,000$ prompts for training and $100$ for testing. The latter contains $750$ prompts for training, $50$ for testing. 
    \item \textbf{Specific task prompts.} Example task prompts for face and hand images.  
\end{itemize}

\para{Reward Functions.} We conduct experiments with different kinds of reward functions as below:
\begin{itemize}
    \item \textbf{Text-to-Image Rewards.} These rewards are trained on text-to-image datasets, such as ImageReward \cite{imagereward} and HPSv2 \cite{hps}.
    \item \textbf{Specific task rewards.} Following \cite{black2023ddpo}, we evaluate model performance for the compression and incompression. Besides, we also design specific rewards for face and hand.
\end{itemize}

Please refer to the supplementary materials for more training details and hyper-parameter settings.

\subsection{Finetuning Text Encoder v.s. U-Net} \label{sec:exp_analysis}

\begin{figure}[!t]
    \centering
    \newcommand{\imgwidth}{0.24\linewidth}
    
    \includegraphics[width=0.8\linewidth]{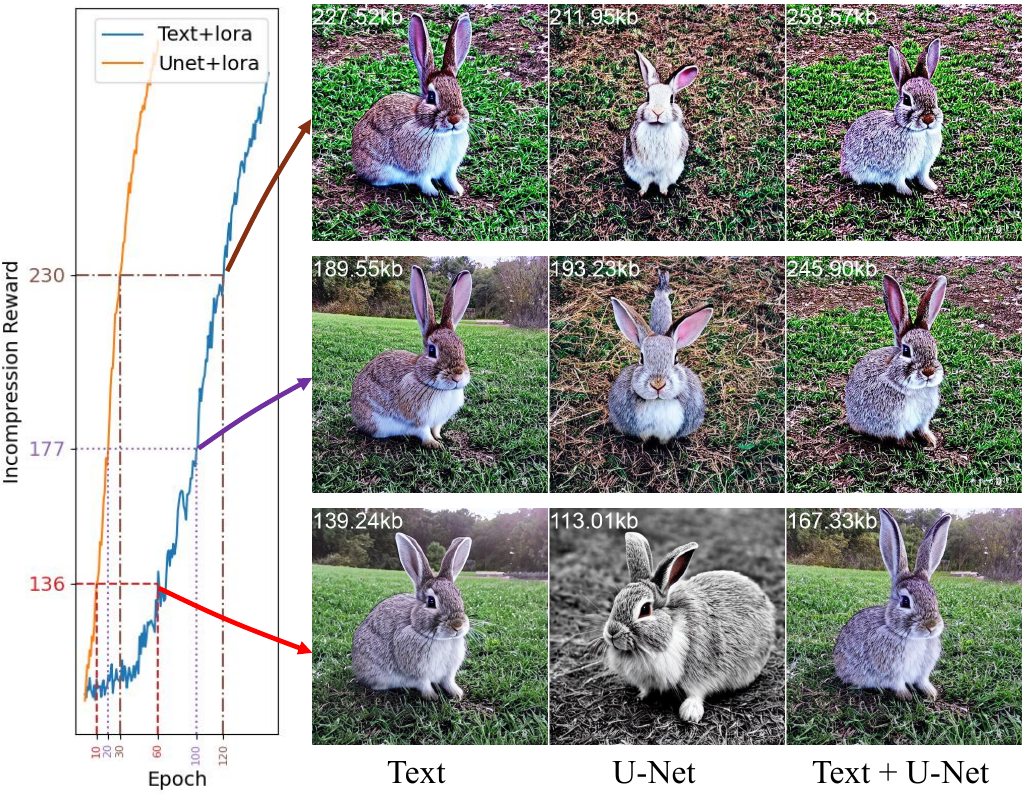}

    \caption{Comparison of training progress between finetuning text encoder and U-Net with LoRA. The image size after JPEG compression is marked on the top-left corner as ``kb''.} \label{fig:compress}
\end{figure}

In this section, we will empirically analyze the difference between finetuning the text encoder and U-Net, through the incompression task as introduced in DDPO \cite{black2023ddpo}. It aims to enhance the complexity of generated images through reinforcement learning (RL). The reward for this task is assessed based on the image size after JPEG compression with a quality factor of $95$. Given its objective nature and the ambiguity of possible solutions, this task is suitable for analyzing behaviors of different models when optimized for the reward. We finetune both the text encoder and the U-Net with LoRA using the simple animal prompts.

\Cref{fig:compress} shows the results comparison between finetuning text encoder and U-Net with LoRA. By comparing models with the same incompression score, we can have the following observations:
\begin{itemize}
    \item U-Net tends to change the visual appearance to increase the reward, whereas the text encoder introduces novel visual concepts to attain the same objective. As shown in \cref{fig:compress}, despite having comparable incompression scores, the outcomes from the text encoder are more coherent than those from the U-Net. However, this also makes the optimization of the text encoder more challenging and time-consuming.
    \item We can directly combine the LoRA weights from TexForce and U-Net to achieve even better results. As shown in \cref{fig:compress}, the results in the third column achieve the highest incompression score and still maintain a similar visual structure from the first column, successfully combining advantages from the LoRA weights of both the text encoder and U-Net. \textit{It is worth noting that this is achieved without additional training.}
\end{itemize}

\subsection{Comparison with Existing Works} \label{sec:exp_compare}

\begin{figure}[!t]
    \centering
    \newcommand{\imgwidth}{0.485\linewidth}
    \newcommand{\imgheight}{0.115\linewidth}

    \textit{\tiny
    \makebox[2pt]{}
    \makebox[\imgheight]{\makecell[c]{A green \\ colored rabbit.}}
    \makebox[\imgheight]{\makecell[c]{A cat \\ and a dog.}}
    \makebox[\imgheight]{\makecell[c]{Four wolves \\ in the park.}}
    \makebox[\imgheight]{\makecell[c]{A dog \\ on the moon.}}
    \makebox[\imgheight]{\makecell[c]{A green \\ colored cat.}}
    \makebox[\imgheight]{\makecell[c]{A cat \\ and a cup.}}
    \makebox[\imgheight]{\makecell[c]{Four birds \\ in the park.}}
    \makebox[\imgheight]{\makecell[c]{A lion \\ on the moon.}}
    }

    \tiny{
    \makebox[5pt][c]{\rotatebox{90}{%
       \makebox[\imgheight][c]{\makecell[c]{DPOK + \\ \textbf{TexForce}}}
       \makebox[\imgheight][c]{\textbf{TexForce}} 
       \makebox[\imgheight][c]{DPOK} 
       \makebox[\imgheight][c]{SDv1.4} 
    }}}
    \includegraphics[width=\imgwidth]{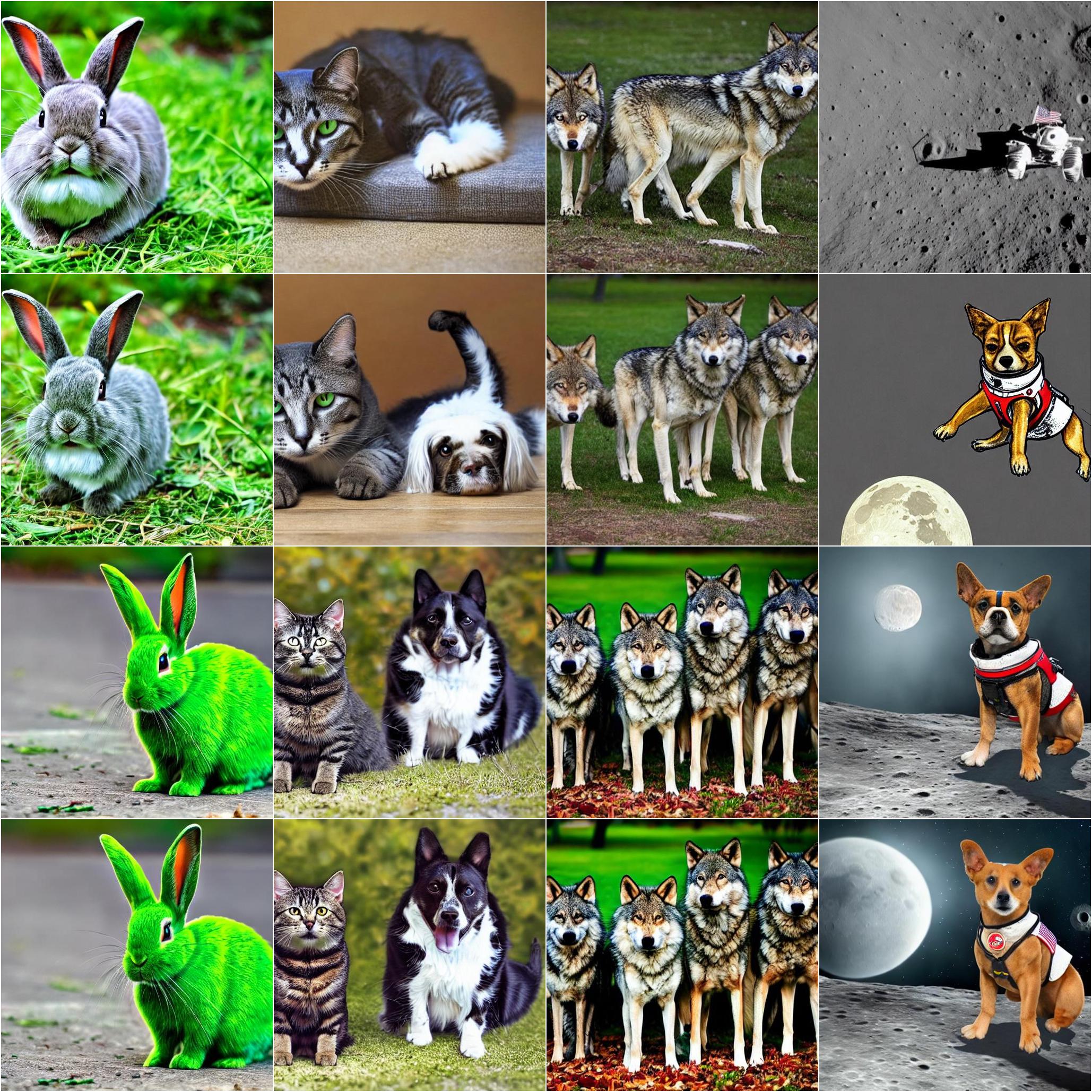}
    \includegraphics[width=\imgwidth]{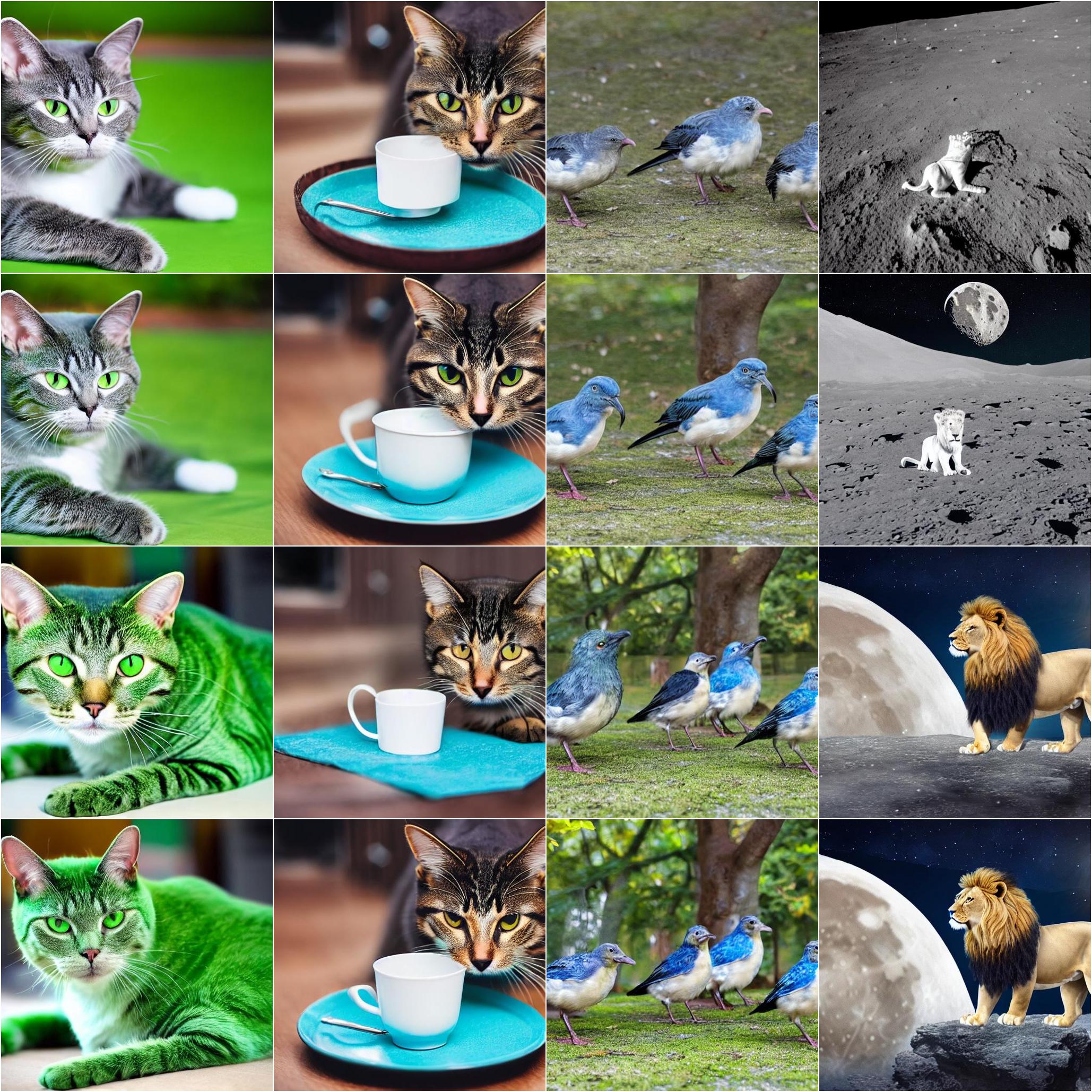}
    
    \includegraphics[width=\imgwidth]{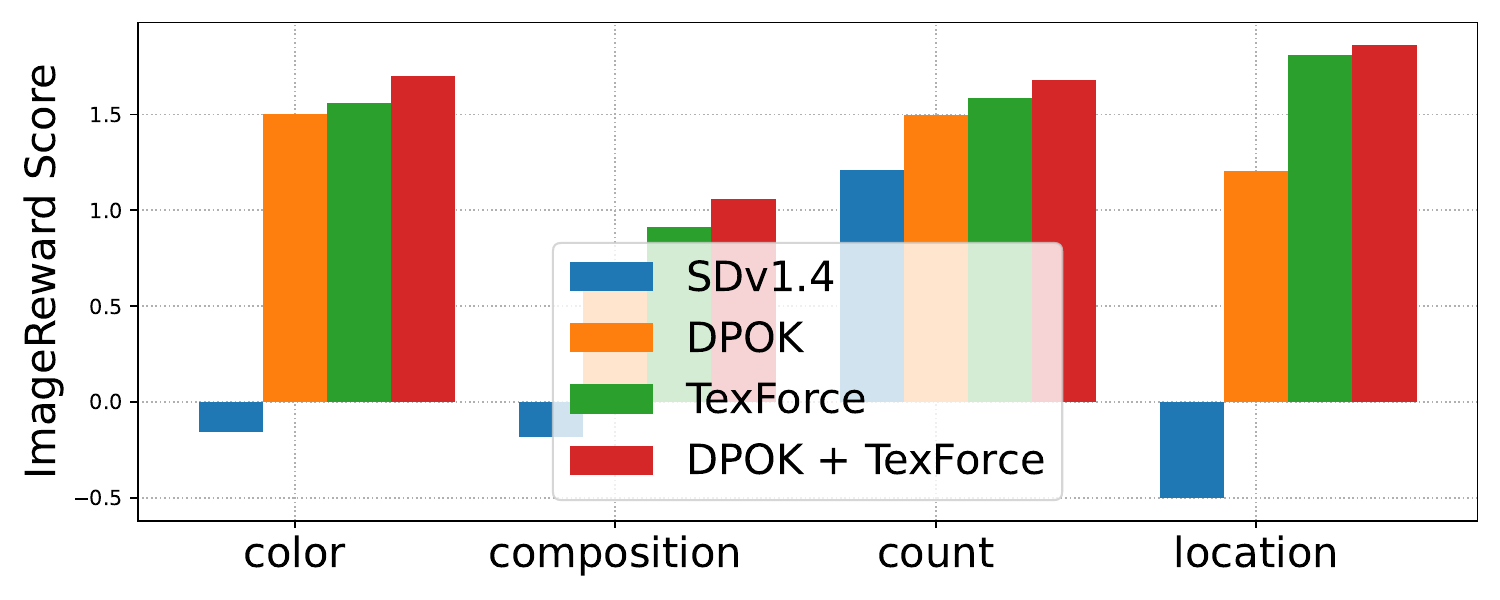}
    \includegraphics[width=\imgwidth]{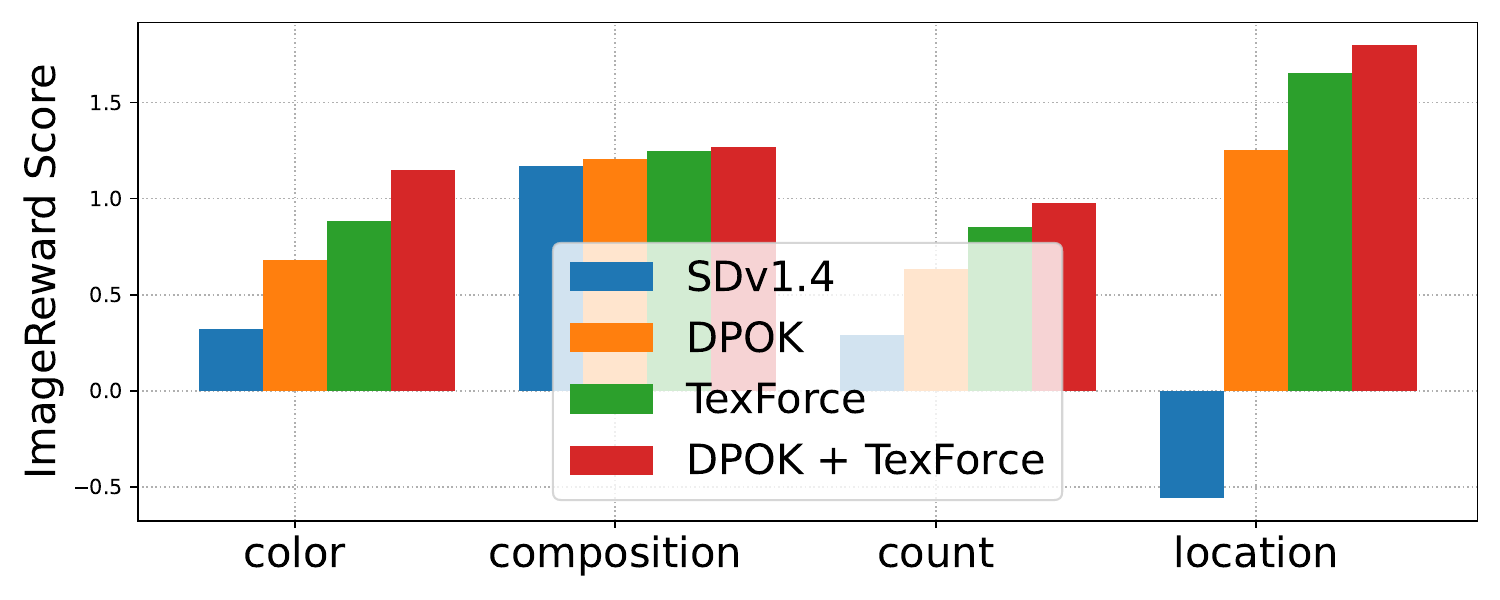}
    \makebox[\imgwidth]{\msmall{(a) Results on seen text prompts}}
    \makebox[\imgwidth]{\msmall{(b) Results on unseen text prompts}}
    \caption{Qualitative and quantitative comparisons with SDv1.4 and DPOK on individual scenarios. Images for comparison are generated with the same random seed. The results show that TexForce can generate more consistent images with better quality than SDv1.4 and DPOK, and simple combination of DPOK and TexForce gives even better performance without any additional training.}
    \label{fig:single_compare_dpok}
\end{figure}

\para{Results on Different Individual Prompts.}
As demonstrated in \cite{lee2023aligning, 2023DPOK}, existing stable diffusion models exhibit misalignment with simple text prompts, such as color consistency (\eg, \textit{A green colored dog.}) and combination of objects (\eg, \textit{A cat and a dog}). Therefore, we start with these simple individual scenarios to highlight the advantages of our method. We follow the experimental settings of DPOK \cite{2023DPOK}, and conduct our experiments with four different capabilities: color consistency, object composition, object count, and object location, as shown in \cref{fig:single_compare_dpok}. Both DPOK and our TexForce are trained with the ImageReward feedback function and 20K samples. All models are evaluated with ImageReward scores and averaged over 50 samples with the same random seeds.

\Cref{fig:single_compare_dpok} presents a comprehensive overview of both quantitative and qualitative results for seen and unseen prompts. The results show that our method can generate more consistent images with better quality than the original SDv1.4 and DPOK. For instance, we can see that the results of TexForce are more consistent with the prompts, such as the color of the rabbit and cat, the number of birds, and the location of the dog. Besides, the results of TexForce are more realistic than DPOK and SDv1.4. Quantitatively, TexForce attains better average ImageReward scores for both seen and unseen prompts. This underscores the overall superiority of TexForce over DPOK and SDv1.4 across multiple samples. Moreover, the combination of TexForce and DPOK demonstrates the best performance in terms of both ImageReward scores and visual quality. This demonstrates the flexibility of our method, which can be seamlessly integrated with existing methods to achieve best performance without additional training.

\begin{figure}[!t]
    \centering
    \small
    \newcommand{\imgheight}{0.19\linewidth}
    \newcommand{\imgwidth}{0.78\linewidth}
    \newcommand{\txtwidth}{0.18\linewidth}

    \msmall{
        \makebox[\imgheight][c]{\textit{Text Prompt}}
        \makebox[\imgheight][c]{SDv1.4} 
        \makebox[\imgheight][c]{ReFL} 
        \makebox[\imgheight][c]{\textbf{TexForce}} 
        \makebox[\imgheight][c]{ReFL+\textbf{TexForce}} 
    }
    
    \begin{minipage}[c]{\txtwidth}
        \scriptsize{Portrait of an old sea captain, male, detailed face, fantasy, highly detailed, cinematic, art painting by greg rutkowski}
    \end{minipage}
    \begin{minipage}[c]{\imgwidth}
        \includegraphics[width=\linewidth]{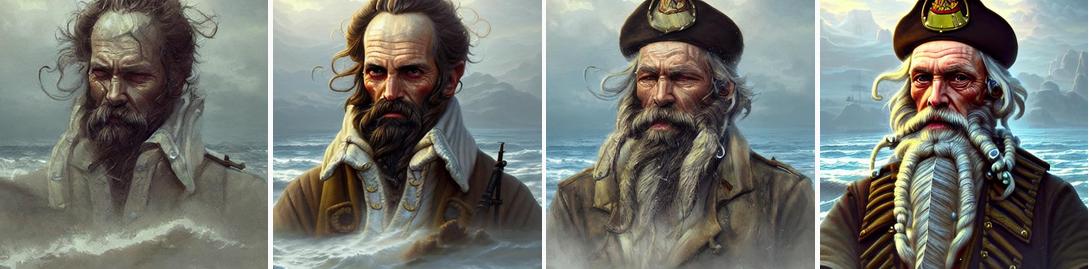}
    \end{minipage}

    \begin{minipage}[c]{\txtwidth}
        \scriptsize{Close up photo of anthropomorphic fox animal dressed in white shirt and khaki cargo pants, fox animal, glasses}
    \end{minipage}
    \begin{minipage}[c]{\imgwidth}
        \includegraphics[width=\linewidth]{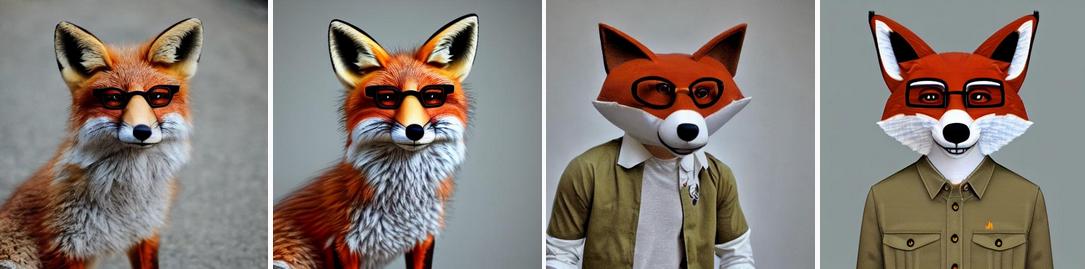}
    \end{minipage}

    \begin{minipage}[c]{\txtwidth}
        \scriptsize{A coffee mug made of cardboard}
    \end{minipage}
    \begin{minipage}[c]{\imgwidth}
        \includegraphics[width=\linewidth]{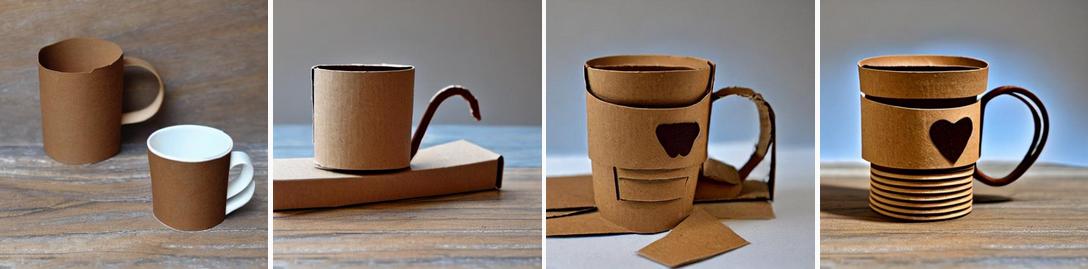}
    \end{minipage}
    
    \caption{Visual comparison with ReFL on ImageReward dataset using real user prompts.} \label{fig:complex_refl}
\end{figure}

\begin{figure}[!t]
    \centering
    \small
    \newcommand{\imgheight}{0.19\linewidth}
    \newcommand{\imgwidth}{0.78\linewidth}
    \newcommand{\txtwidth}{0.18\linewidth}

    \msmall{
        \makebox[\imgheight][c]{\textit{Text Prompt}}
        \makebox[\imgheight][c]{SDv1.4} 
        \makebox[\imgheight][c]{AlignProb} 
        \makebox[\imgheight][c]{\textbf{TexForce}} 
        \makebox[\imgheight][c]{AlignProb+\textbf{TexForce}} 
    }
    
    \begin{minipage}[c]{\txtwidth}
        \scriptsize{
        Bicycles with back packs parked in a public place.
        }
    \end{minipage}
    \begin{minipage}[c]{\imgwidth}
        \includegraphics[width=\linewidth]{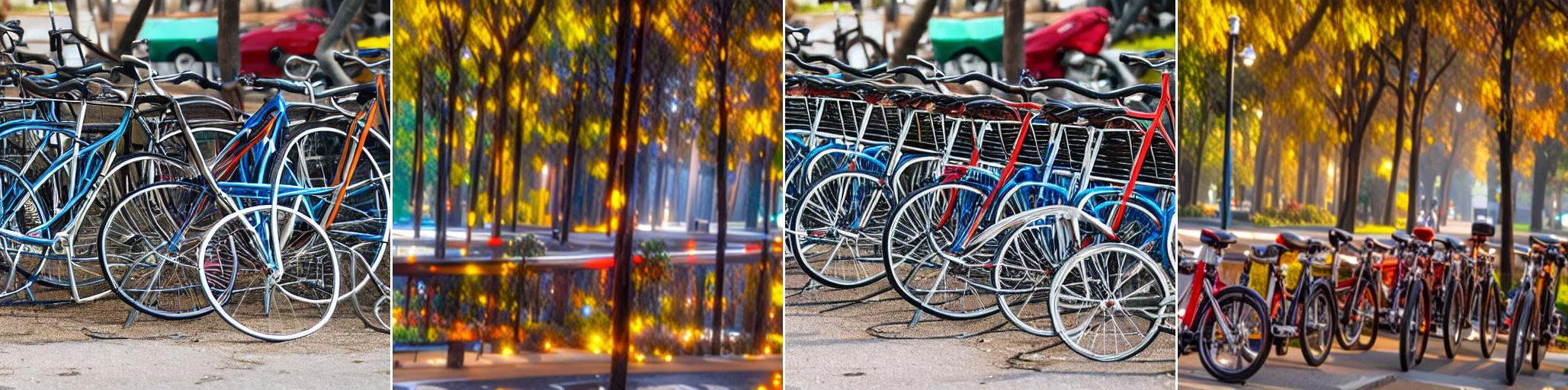}
    \end{minipage}

    \begin{minipage}[c]{\txtwidth}
        \scriptsize{A dog with a plate of food on the ground
        }
    \end{minipage}
    \begin{minipage}[c]{\imgwidth}
        \includegraphics[width=\linewidth]{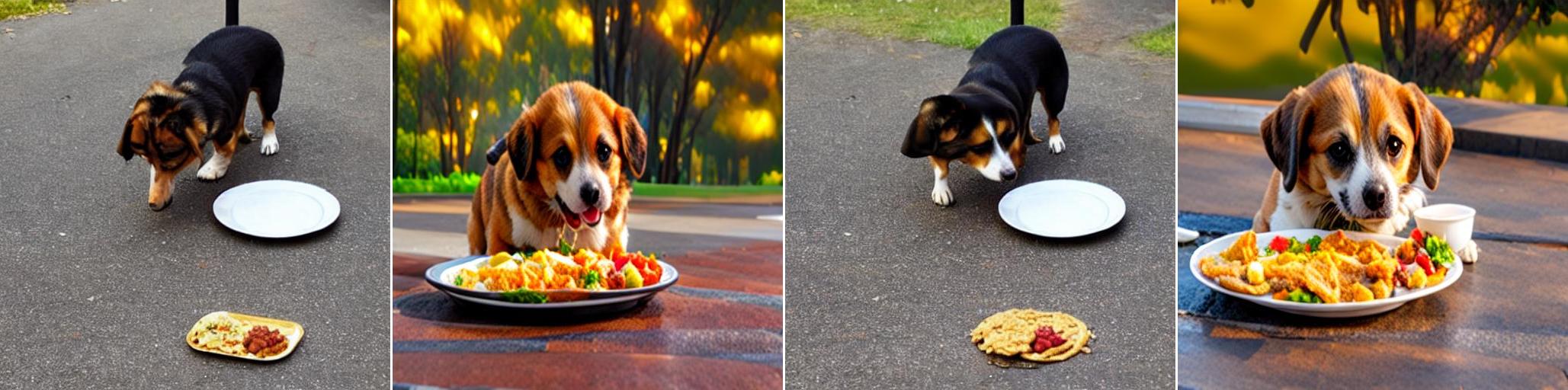}
    \end{minipage}

    \begin{minipage}[c]{\txtwidth}
        \scriptsize{A large commercial airliner silhoetted in the sun}
    \end{minipage}
    \begin{minipage}[c]{\imgwidth}
        \includegraphics[width=\linewidth]{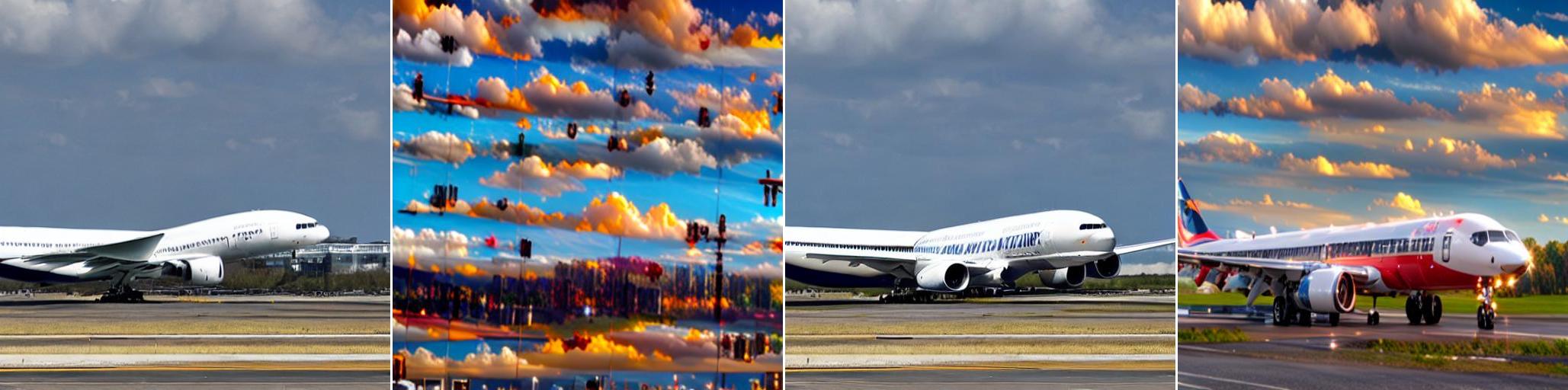}
    \end{minipage}
    
    \caption{Visual comparison with AlignProb on HPSv2 dataset using real user prompts.} \label{fig:complex_alignprop}
\end{figure}

\para{Results on Complex Long Prompts} Next, we conduct experiments using larger dataset with complex long prompts, \ie, the ImageReward dataset \cite{imagereward} and HPS dataset \cite{hps}. We retrained with ReFL and AlignProb with official codes, and the results are shown in \cref{fig:complex_refl,fig:complex_alignprop}. As we can observe, since ReFL is trained with a single step backward to update the U-Net, it is less effective than the proposed TexForce to align the input prompts with generated images, such as the white shirt of the fox. The improvement of ReFL is mainly the appearance of the generated images, such as the color of the man and the texture of the fox. Meanwhile, the proposed TexForce is better at aligning the text prompts with generated images, which makes TexForce better in \cref{tab:complex}. Furthermore, when merging the strengths of TexForce and ReFL, we observe a notable improvement in both quantitative results and visual appearance.

\begin{table}[!t]
    \caption{Quantitative results on ImageReward and HPSv2. Results are tested with the same seed and prompts.}
    \label{tab:complex}
    \begin{subtable}[t]{0.48\linewidth}
        \centering
        \setlength{\tabcolsep}{6pt}
        \begin{tabular}{c | c}
        \hline
        Method & ImageReward \\
        \hline 
        SDv1.4 & 0.2154 \\
        ReFL & 0.4485 \\
        TexForce & \second{0.4556} \\
        ReFL + TexForce & \best{0.6553} \\
        \hline
       \end{tabular}
    \end{subtable}
    \begin{subtable}[t]{0.48\linewidth}
        \centering
        \setlength{\tabcolsep}{6pt}
        \begin{tabular}{c | c }
        \hline
        Method & HPSv2  \\
        \hline
        SDv1.4 & 0.2752 \\
        AlignProb & \second{0.2821} \\
        TexForce & 0.2767 \\
        AlignProb + TexForce & \best{0.2914} \\
        \hline
        \end{tabular}
     \end{subtable}
\end{table}

Similarly, we compared our method with AlignProb using the HPSv2 reward. The results from AlignProb appeared overly optimized towards the rewards, evident in the abundance of yellow spotlights and clouds, leading to disrupted semantics. In contrast, our proposed TexForce primarily aims to improve the text-image alignment. While our reward score was slightly lower than AlignProb due to limited changes in color style, our results better match the textual prompts in terms of semantics. Additionally, our combined results maintain the visual styles preferred by the HPSv2 while preserving the meaning of the text prompts, resulting in significant improvement over AlignProb.

\subsection{Experiments with Different Backbones} \label{sec:backbone}

\begin{figure}[!t]
    \centering
    \small
    \newcommand{\imgheight}{0.19\linewidth}
    \newcommand{\imgwidth}{0.78\linewidth}
    \newcommand{\txtwidth}{0.18\linewidth}

    \msmall{
        \makebox[\imgheight][c]{\textit{Text Prompt}}
        \makebox[\imgheight][c]{SDv1.5} 
        \makebox[\imgheight][c]{ReFL} 
        \makebox[\imgheight][c]{\textbf{TexForce}} 
        \makebox[\imgheight][c]{ReFL+\textbf{TexForce}} 
    }
    
    \begin{minipage}[c]{\txtwidth}
        \scriptsize{
        Victorian lady, painting by rossetti, daniel gerhartz, alphonse mucha, bouguereau, detailed art
        }
    \end{minipage}
    \begin{minipage}[c]{\imgwidth}
        \includegraphics[width=\linewidth]{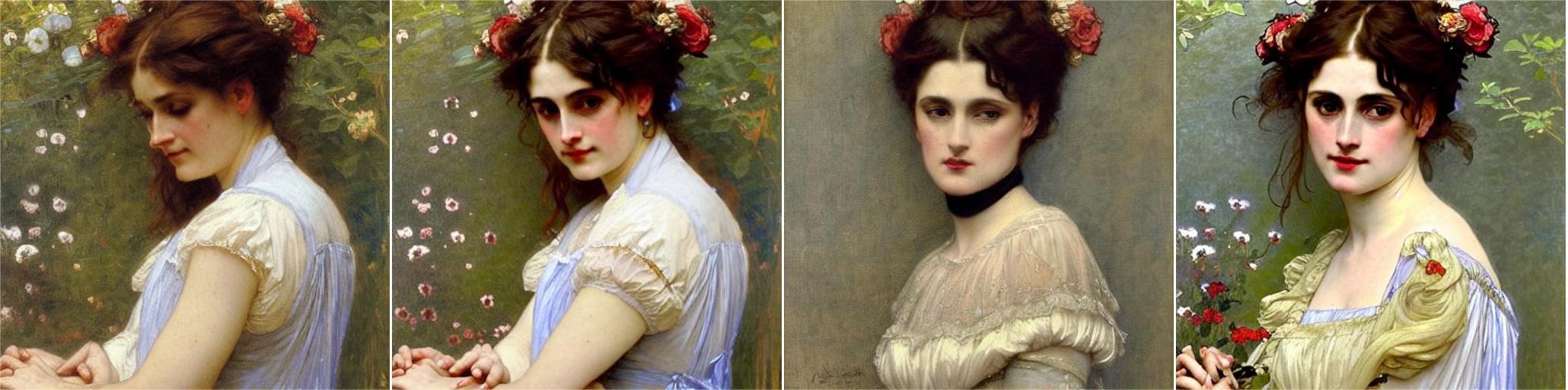}
    \end{minipage}

    \begin{minipage}[c]{\txtwidth}
        \scriptsize{
        medieval old king, character, hearthstone, fantasy, elegant, highly, illustration, art by artgerm and greg rutkowski and alphonse much
        }
    \end{minipage}
    \begin{minipage}[c]{\imgwidth}
        \includegraphics[width=\linewidth]{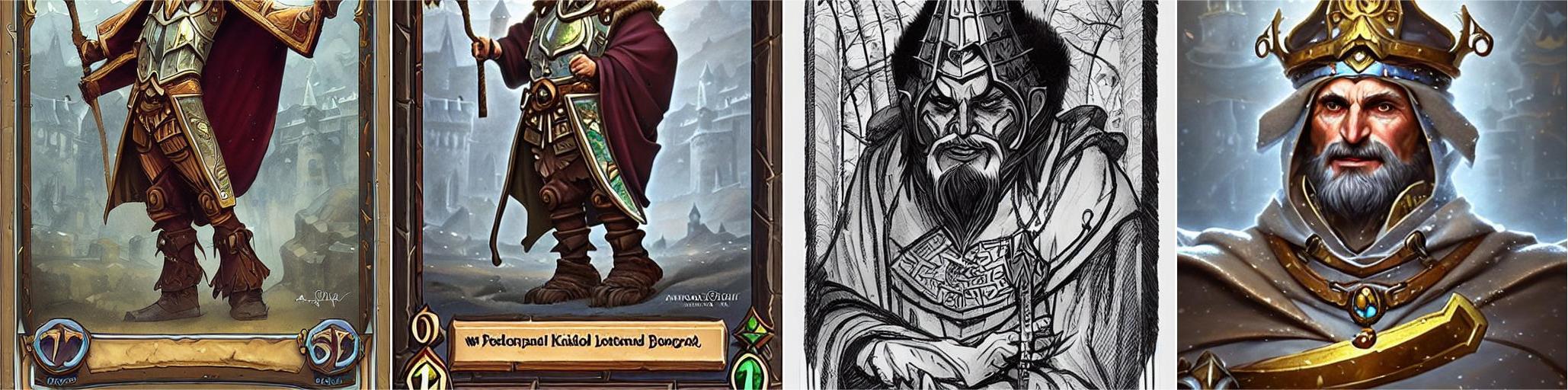}
    \end{minipage}

    \begin{minipage}[c]{\txtwidth}
        \scriptsize{
        Classic model of atoms, made out of glass marbles and chrome steel rods, studio
        }
    \end{minipage}
    \begin{minipage}[c]{\imgwidth}
        \includegraphics[width=\linewidth]{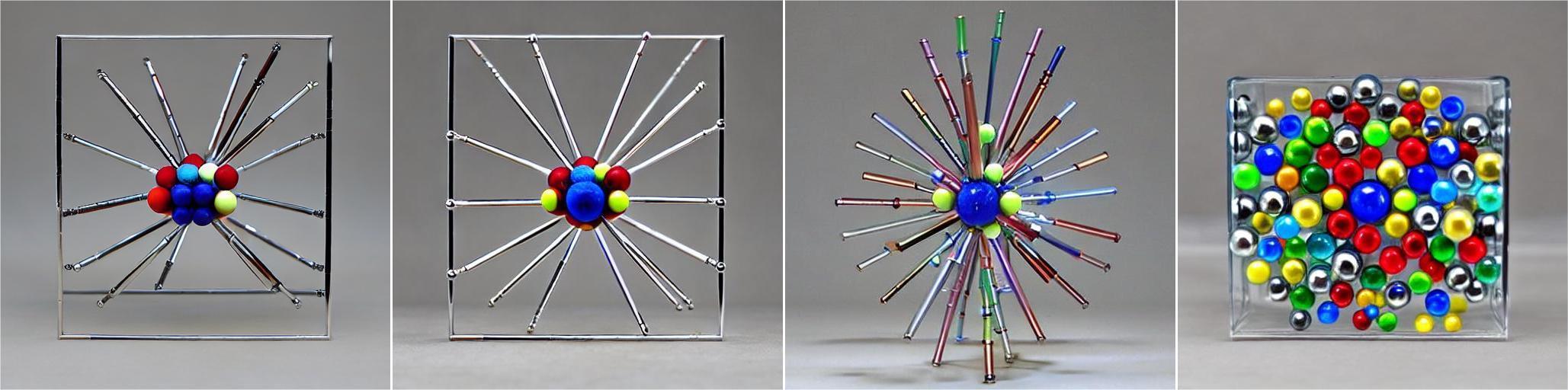}
    \end{minipage}

    \vspace{3pt}
    \par\noindent\rule{\textwidth}{0.6pt}
    
    \msmall{
        \makebox[\imgheight][c]{\textit{Text Prompt}}
        \makebox[\imgheight][c]{SDv2.1} 
        \makebox[\imgheight][c]{ReFL} 
        \makebox[\imgheight][c]{\textbf{TexForce}} 
        \makebox[\imgheight][c]{ReFL+\textbf{TexForce}} 
    }
    
    \begin{minipage}[c]{\txtwidth}
        \scriptsize{
        Victorian lady, painting by rossetti, daniel gerhartz, alphonse mucha, bouguereau, detailed art
        }
    \end{minipage}
    \begin{minipage}[c]{\imgwidth}
        \includegraphics[width=\linewidth]{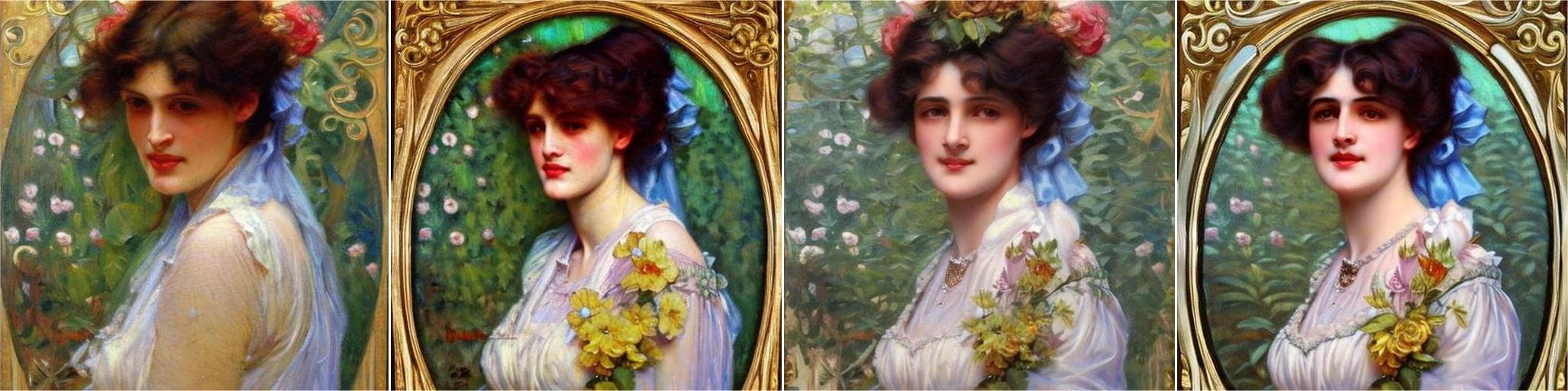}
    \end{minipage}

    \begin{minipage}[c]{\txtwidth}
        \scriptsize{
        medieval old king, character, hearthstone, fantasy, elegant, highly, illustration, art by artgerm and greg rutkowski and alphonse much
        }
    \end{minipage}
    \begin{minipage}[c]{\imgwidth}
        \includegraphics[width=\linewidth]{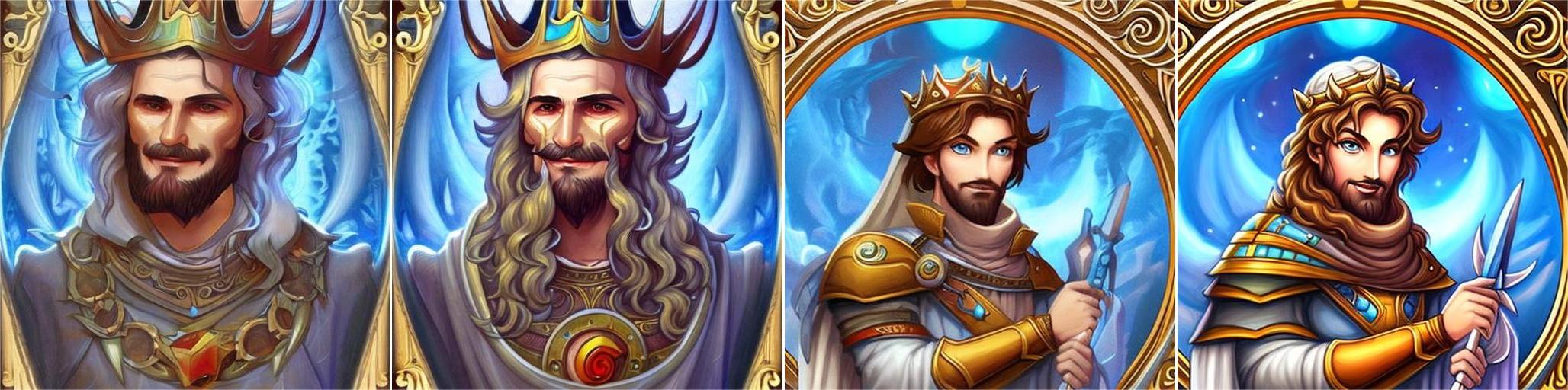}
    \end{minipage}

    \begin{minipage}[c]{\txtwidth}
        \scriptsize{
        Classic model of atoms, made out of glass marbles and chrome steel rods, studio
        }
    \end{minipage}
    \begin{minipage}[c]{\imgwidth}
        \includegraphics[width=\linewidth]{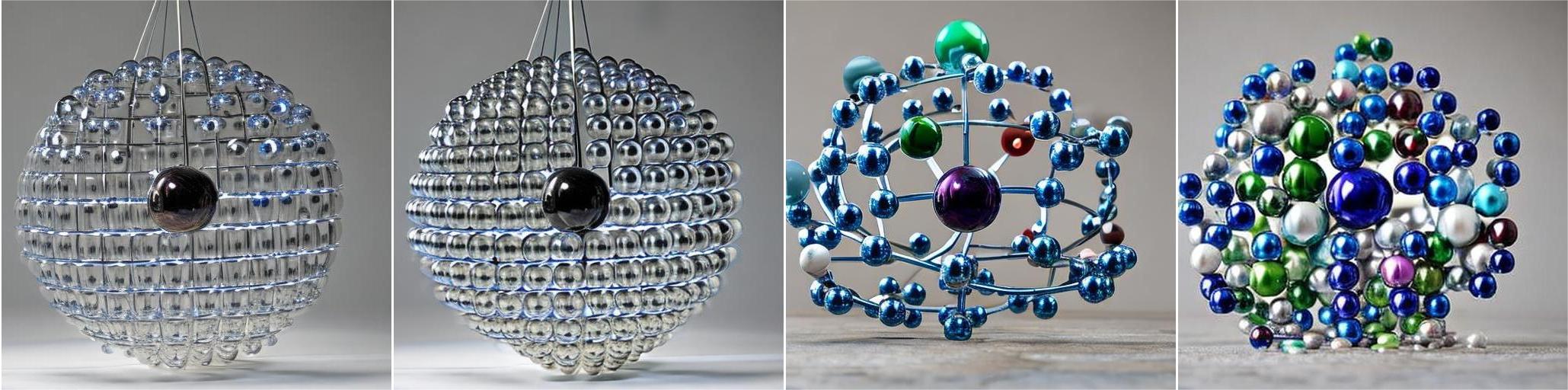}
    \end{minipage}
    
    \caption{Results with SDv1.5 and SDv2.1 backbones on ImageReward test dataset.} \label{fig:backbone} 
\end{figure}

To show the robustness of our method, we conduct experiments with more different backbones, including SDv1.5\footnote{\url{https://huggingface.co/runwayml/stable-diffusion-v1-5runwayml/stable-diffusion-v1-5}} and SDv2.1\footnote{\url{https://huggingface.co/stabilityai/stable-diffusion-2-1}}. We use the ImageReward score and prompts dataset to train all the models, and the results are shown in \cref{fig:backbone} and \cref{tab:backbone}. Notably, our method consistently improves the text-image alignment of the original models. For instance, the results generated by TexForce exhibit enhanced visual appeal and better consistency with the prompts, such as the \textit{victorian lady}, \textit{old king}, and \textit{atom model}. It is also worth noting that although SDv2.1 is already much better than SDv1.5, TexForce continues to augment the performance. This demonstrates the adaptability and robustness of our method when employed with different backbones. Although ReFL achieves higher ImageReward scores, our observations indicate that it primarily enhances color and fine details and is less effective than TexForce in aligning images with text prompts. For both SDv1.5 and SDv2.1, the combined model yields the best performance, which clearly affirms the effectiveness of TexForce.  

\begin{table}[!t]
    \centering
    \caption{Quantitative results with SDv1.5 and SDv2.1 on the ImageReward dataset.} \label{tab:backbone}
    \setlength{\tabcolsep}{6pt}
    \begin{tabular}[\linewidth]{c|cccc}
    \hline
        Backbone & Original & ReFL & TexForce & ReFL+TexForce \\ \hline 
        SDv1.5 & 0.2140 & 0.5484 & 0.4086 & \best{0.6703} \\
        SDv2.1 & 0.3891 & 0.5223 & 0.5084 & \best{0.6158} \\
    \hline
    \end{tabular}
\end{table}

\subsection{GPT-4V Evaluation}

\begin{figure}[!t]
    \centering
    \includegraphics[width=0.9\linewidth]{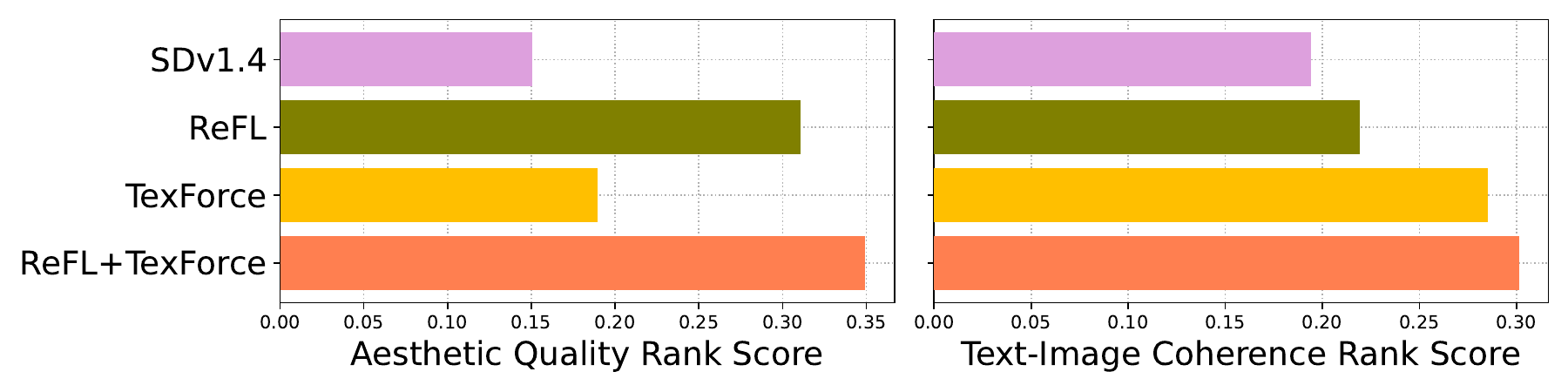}
    \caption{GPT4V evaluation for aesthetic quality and text-image coherence with ImageReward testset and SDv1.4. Refer to supplementary for SDv1.5 and SDv2.1 results.}
    \label{fig:gpt4v}
\end{figure}

As GPT-4V has recently shown to be comparable with human-level performance in evaluating image quality \cite{wu2023qbench,wu2023qinstruct}, we decide to rely on GPT-4V evaluations instead of traditional user studies, which may be inconsistent and hard to reproduce. Our approach involves using GPT-4V to rank image quality based on aesthetic quality and coherence with text. In \cref{fig:gpt4v}, we present the average scores from three rounds of evaluations using the ImageReward test dataset. We can see that our TexForce method does a better job at aligning text with images in diffusion models, while ReFL improves the appearance of the images. Combining both approaches successfully takes advantage of both of them and yields the best results. Please refer to the supplementary material to reproduce the results.

\subsection{Ablation Study about Joint Finetune}

\begin{table}[t]
    \centering
    \setlength{\tabcolsep}{6pt}
    \caption{Quantitative comparison between simple fusion and joint training.} \label{tab:joint}
    \begin{tabular}{c|cccc|c}
        \hline
        Methods & SDv1.4 & ReFL & TexForce & ReFL + TexForce & Joint \\ \hline
        Score & 0.2154 & 0.4485 & 0.4556 & \best{0.6553} & \second{0.5009} \\ \hline 
    \end{tabular}
\end{table}

\begin{figure}[t]
    \newcommand{\imgwidth}{0.19\linewidth}
    
    \begin{minipage}{\textwidth}
        \centering
        \scriptsize{a beautiful cyborg mermaid with a long fish tail, the body has shimmering fish scales, submerged underwater, dark ocean, light filtering through, the body is entwined in seaweed and coral, highly detailed, hyper - realistic, futuristic}
    \end{minipage}   
    \includegraphics[width=\linewidth]{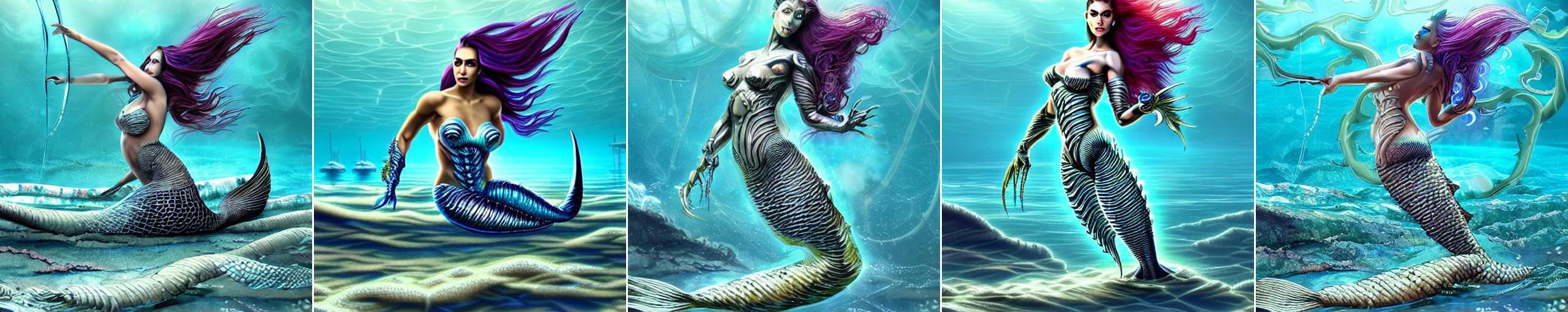}
    \begin{minipage}{\textwidth}
        \centering
        \scriptsize{footage of an astronaut in a tropical beach}
    \end{minipage}   
    \includegraphics[width=\linewidth]{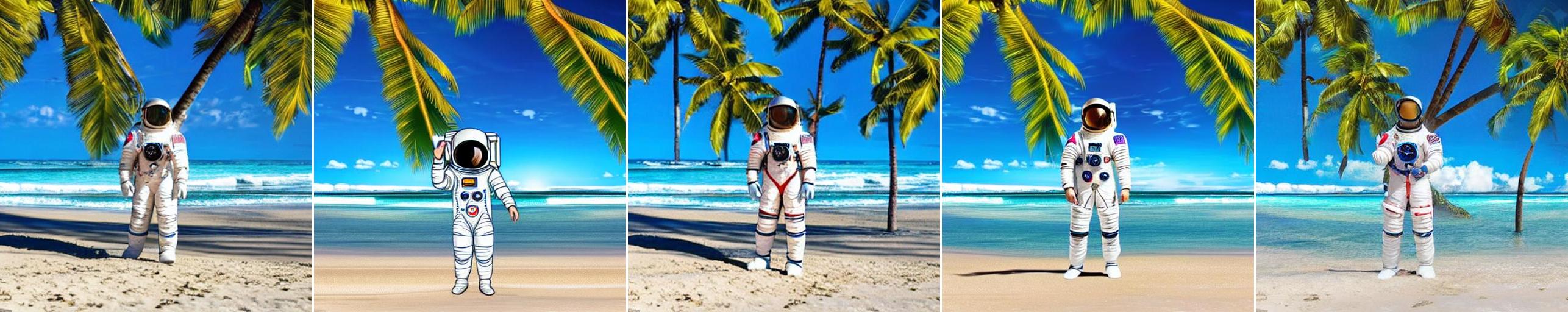}
    \makebox[\imgwidth]{SDv1.4}
    \makebox[\imgwidth]{ReFL}
    \makebox[\imgwidth]{TexForce}
    \makebox[\imgwidth]{ReFL+TexForce}
    \makebox[\imgwidth]{Joint}

    \caption{Visual comparison between simple fusion and joint training.} \label{fig:joint}
\end{figure}

Given the efficacy of straightforward fusion, it prompts us to inquire whether a joint finetuning approach yields good results. We conducted such experiments using the SDv1.4 backbone, and the results are illustrated in \cref{tab:joint} and \cref{fig:joint}. It is evident that, although the quantitative performance of joint finetuning surpasses that of ReFL and TexForce, it still falls short of the performance achieved by their simple combination. We hypothesize that this disparity arises because the fixed U-Net can serve as a prior for pixel generation during the finetuning of the text encoder. Consequently, joint fine-tuning complicates the optimization process for the text encoder, thereby leading to inferior results.

\subsection{Applications} \label{sec:exp_applications}

\begin{figure}[!t]
    \begin{minipage}[c]{.46\textwidth}
        \includegraphics[width=\linewidth]{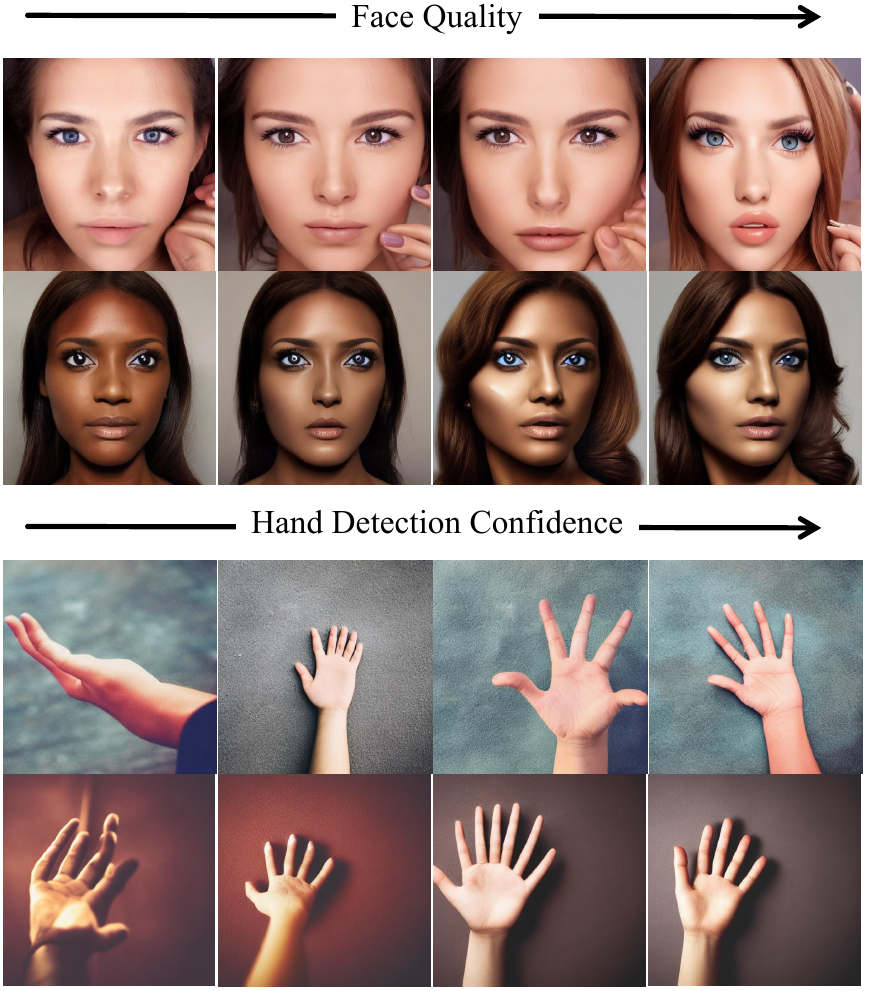}
        \caption{Two example applications of TexForce: high-quality face and hand generation.} \label{fig:face}
    \end{minipage}
    \hfill
    \begin{minipage}[c]{.53\textwidth}
        \includegraphics[width=\linewidth]{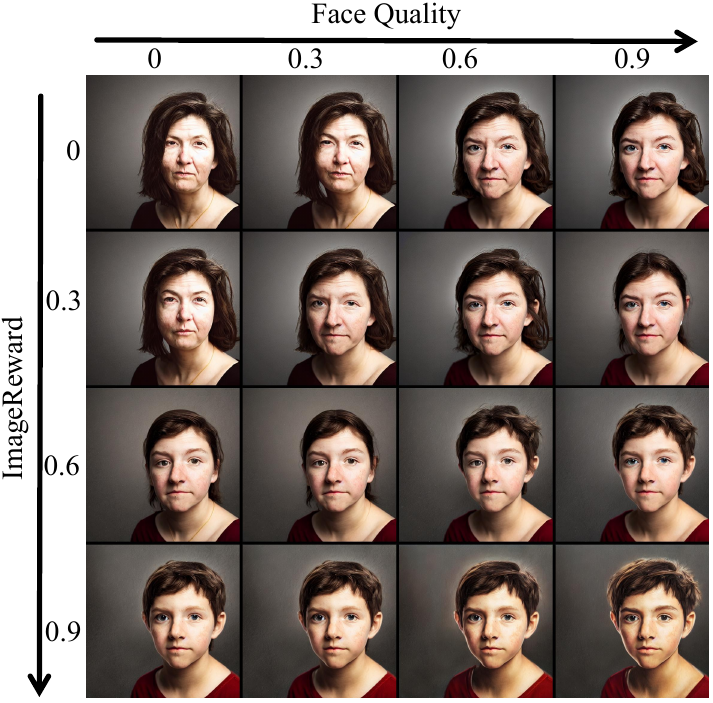}
        \caption{Fusion of ImageReward LoRA weights and face quality LoRA weights. Prompt: \emph{A realistic portrait photo.}} \label{fig:fusion}
    \end{minipage}
\end{figure}

TexForce demonstrates remarkable adaptability to diverse tasks, as it does not require differentiable rewards. In this section, we showcase its capabilities in enhancing the quality of generated face and hand images.

\para{Face reward.} We employ the face quality evaluation metric from \cite{pyiqa}, which is based on an image quality evaluation network \cite{chen2023topiq} trained using the face quality dataset \cite{gfiqa20k}.

\para{Hand reward.} Regarding the hand quality evaluation, we recognize the absence of specific hand quality metrics. Instead, we employ a straightforward hand detection confidence score as a reward function and observe its utility. The hand detection model from \cite{Alam2021Unified} is used to calculate the confidence score.

\Cref{fig:face} illustrates the progressive improvement in the quality of generated face and hand images over the course of training. These results illustrate the capacity of TexForce to enhance image quality, utilizing either direct quality metrics or a simple confidence score.

Moreover, by utilizing LoRA weights for fine-tuning the text encoder, we find that it is feasible to blend specific LoRA weights to enhance the quality of specific objects. Suppose the LoRA weight $\theta_i$ from $i$-th task, we can simply fuse them via $\sum_i\alpha_i\theta_i$. In \cref{fig:fusion}, we demonstrate how the fusion of ImageReward LoRA weights and face quality LoRA weights can produce high-quality face images. This flexibility significantly broadens the range of potential applications for TexForce.

\section{Conclusion}

In this paper, we introduce a new method called \textbf{TexForce} for enhancing the text encoder of diffusion models using reinforcement learning. Our research demonstrates that refining the text encoder can enhance the overall performance of diffusion models, specifically in terms of aligning text and images as well as improving visual quality. Furthermore, we illustrate that TexForce can be seamlessly integrated with existing U-Net models that have undergone fine-tuning, without the need for extra training, resulting in significant performance improvements. Lastly, we showcase the versatility of our approach across various applications, including the generation of high-quality images of faces and hands. We also provide evidence that the finetuned LoRA weights with different tasks can be combined to enhance the specific quality of image generation.

\section*{Acknowledgments} 
This study is supported under the RIE2020 Industry Alignment Fund – Industry Collaboration Projects (IAF-ICP) Funding Initiative, as well as cash and in-kind contribution from the industry partner(s).

{
    \bibliographystyle{splncs04}
    \bibliography{main}
}

\newpage



\section*{Supplementary material}

# A  Method Details

## A.1  Reinforcement Learning Formulations

We first give the formulations of each term for the PPO algorithm in diffusion models. According to the DDPM paper [16], the $t$ step denoising can be formulated as:

$$\mathbf{x}_{t-1} = \mu_{\theta,\phi}(\mathbf{x}_t, t, \tau_\phi(s)) + \sigma_t \epsilon, \tag{7}$$

$$\mu_{\theta,\phi} = \frac{1}{\sqrt{\alpha_t}} \left( \mathbf{x}_t - \frac{\beta_t}{\sqrt{1-\bar{\alpha}_t}} \epsilon_{\theta,\phi}(\mathbf{x}_t, t, \tau_\phi(s)) \right) \tag{8}$$

where $\bar{\alpha}_t = \prod_{i=1}^{t} \alpha_i$, $\sigma_t^2 = \frac{1-\bar{\alpha}_{t-1}}{1-\bar{\alpha}_t}\beta_t$, $\epsilon \sim \mathcal{N}(\mathbf{0}, \mathbf{I})$, and $\beta_t$ is a predefined variance schedule. Since the U-Net parameter $\phi$ is fixed during the training, we omit it in the following formulations. The objective function of the PPO algorithm is:

$$J(\phi) = \mathbb{E}\left[\min(r_t(\phi)A, \text{clip}(r_t(\phi), 1-\lambda, 1+\lambda)A)\right], \tag{9}$$

where

$$r_t(\phi) = \frac{p_\phi(\mathbf{x}_{t-1}|\mathbf{x}_t)}{p_{\phi_{\text{old}}}(\mathbf{x}_{t-1}|\mathbf{x}_t)}. \tag{10}$$

Since $p_\phi$ is an isotropic Gaussian distribution, we have:

$$\begin{aligned} r_t(\phi) &= \exp(\log p_\phi(\mathbf{x}_{t-1}|\mathbf{x}_t) - \log p_{\phi_{\text{old}}}(\mathbf{x}_{t-1}|\mathbf{x}_t)) \\ &= \exp\left(-\frac{1}{2\sigma_t^2}\|\mathbf{x}_{t-1} - \mu_{\phi,t}\|^2 + \frac{1}{2\sigma_t^2}\|\mathbf{x}_{t-1} - \mu_{\phi_{\text{old}},t}\|^2\right). \end{aligned} \tag{11}$$

The reward value $A$ is obtained with reward function $A = R(\mathbf{x}_0, s)$. With equations above, we can get $J(\phi)$.

## A.2  Face and Hand Rewards

Figure 13 shows the pipeline of face reward and hand reward. For the face quality reward, we finetuned the TOPIQ model[5] [5] with the GFIQA-20k dataset [46] which is specifically designed for face quality. Since the faces of GFIQA are all aligned, we also need to align the generated face before calculating the reward scores. For hand reward function, there is no existing quality model for hands. We found that simple hand detection confidence score can already give reliable reward for the generation quality of hands. Therefore, we directly use the hand detection confidence as rewards. Thanks to the flexibility of RL, the reward function is not required to be differentiable. We use the pretrained YOLOv5[6] for hand detection.

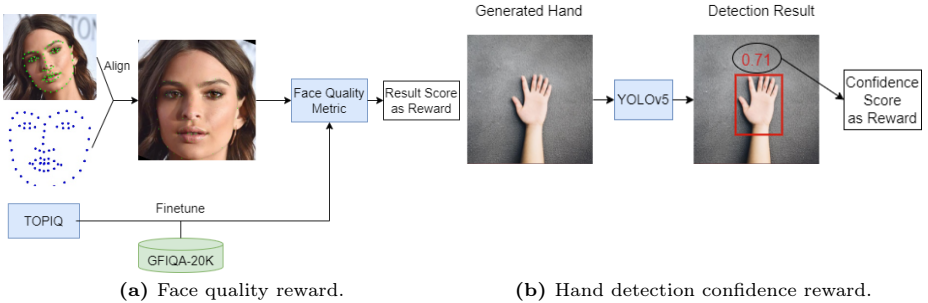

(a) Face quality reward.     (b) Hand detection confidence reward.

**Fig. 13:** Calculation of face quality and hand detection reward.

**Table 4:** Hyper-parameters and training settings for single prompt and multi-prompts datasets.

|  | Hyper-parameters | Single prompt | Multiple prompts |
|---|---|---|---|
| Diffusion | Sampler | DDIM [43] | |
|  | Guidance Scale | 7.5 | |
|  | Sampling Steps | 50 | |
| Optimizer | Type | AdamW | |
|  | Learning rate | 3e-4 | |
|  | Weight decay | 1e-4 | |
|  | $(\beta_1, \beta_2)$ | (0.9, 0.999) | |
|  | Gradient clip | 1.0 | |
| RL Config | Ratio clip ($\gamma$) | 1e-4 | |
|  | Advantage clip | 10 | |
| LoRA | Rank | 16 | |
|  | Alpha ($\alpha$) | 1 | |
|  | Module | q, k, v, out | |
| Training | Trainable #Params. | 1.18M | |
|  | Numerical precision | torch.float16 | |
|  | Batch size | 8 | 64 |
|  | Samples per epoch | 256 | 2048 |
|  | Total epochs | 100 | 100 |
|  | GPUs | 1 V100 | 4 A100 |
|  | Time | ~2 days | ~3 days |

# B Experiment Settings

## B.1 Training Details

As shown in Tab. 4, we use the same hyper-parameters for all the experiments. The only difference is the batch size and the number of samples per epoch. For the single prompt dataset, we use a batch size of 8 and 256 samples per epoch. For the multi-prompts dataset, we use a batch size of 64 and 2048 samples per epoch.

## B.2 Evaluation Details

We use the `seed_everything()` function provided by PyTorch Lightning[7] to set the random seed for all the experiments. **We set the start random seed to 234 for ALL prompts.**

For the single prompt dataset, we generate 50 examples for each prompt. For the multi-prompts dataset, ImageReward and HPSv2, we generate only one example for each prompt to save time.

# C GPT4V Evaluation

We use the `gpt-4-1106-vision-preview` API[8] provided by OpenAI to evaluate the quality of the generated images. The API takes a text prompt and a list of images as input, and returns a rank of the image names based on their aesthetic quality and their coherence with the prompt. Below is an example we used for the evaluation:

---

[5] https://github.com/chaofengc/IQA-PyTorch
[6] https://github.com/MahmudulAlam/Unified-Gesture-and-Fingertip-Detection
[7] https://lightning.ai/
[8] https://platform.openai.com/docs/guides/vision

**GPT4V Evaluation Example (gpt-4-1106-vision-preview)**

**Prompt (#User):**
The following four images, Image 1, Image 2, Image 3 and Image 4, are generated with the prompt [*alien landscape with futuristic portal to another alien planet, astronaut stepping through the portal*], please rank them based on their aesthetic quality and their coherence with the given prompt. Answer with only two lists with the image file names, from good to bad. Please only give two lists.

Image 1(SDv1.4)    Image 2(ReFL)    Image 3(TexForce)   Image 4(ReFL+TexForce)

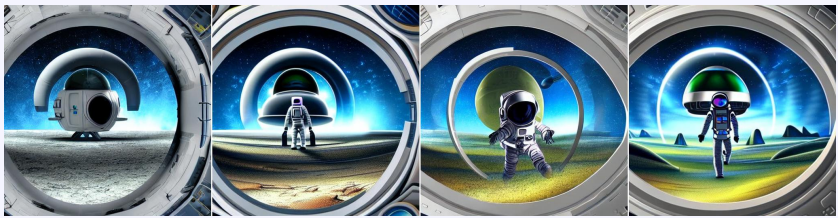

**GPT4V Output:**
Aesthetic quality, from good to bad:
- Image 2      # ReFL
- Image 4      # ReFL+TexForce
- Image 3      # TexForce
- Image 1      # SDv1.4

Coherence with prompt, from good to bad:
- Image 4      # ReFL+TexForce
- Image 3      # TexForce
- Image 2      # ReFL
- Image 1      # SDv1.4

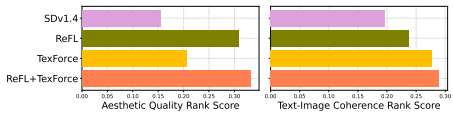
(a) Round 1

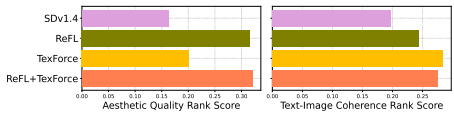
(b) Round 2

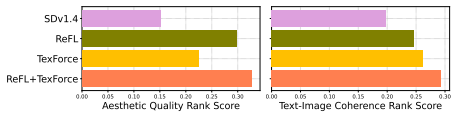
(c) Round 3

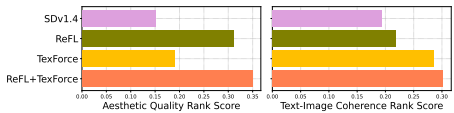
(d) Average

**Fig. 14:** Complete results of GPT4V evaluation with SDv1.4 backbone on the ImageReward dataset.

As shown in the example above, GPT4V returns two lists of the image names, ranking from good to bad. We assign score 3 for the best image, 0 for the worst image and the final score is normalized to [0, 1]. For reliability, we run the evaluation for 3 times and report the average score. The results for each round and the final test score is reported in Fig. 14.

## D  Additional Experiments

### D.1  More Results of Incompression Task

In the main paper, we briefly discussed different behaviors of finetuing U-Net and text encoder in the incompression task. Here, we provide more quantitative and qualitative results about the comparison between finetuning U-Net and text encoder, with the unseen animal prompts below:

**Animal prompts.** Following previous works, we use the 45 simple animals for training. The prompt is defined as *A photo of a <animal>*. We use the following animals for testing: *cheetach, elephant, girraffe, hippo, jellyfish, panda, penguin, swan*.

We generate 10 samples for each prompt and report their incompression scores in Tab. 5. We can have the following observations:

– Although the training rewards of text encoder and U-Net are similar, the text encoder is more robust in unseen animals than U-Net, and obtained higher incompression scores.
– The quantitative results also confirm that simply combining U-Net and text encoder is quite effective in improving the reward scores.

Figure 15 shows visual examples of combining U-Net and text encoder at different checkpoints. We can observe that the text encoder can introduce extra reasonable visual concepts to the image to increase complexity, however, U-Net mainly changes the appearance and is easy to disrupt the original structure such as the hippo head.

**Table 5:** Quantitative results of combining U-Net and text encoder in the incompression task. The comparison checkpoints are the epochs for U-Net and text encoder to achieve the same evaluation reward.

| Comparison checkpoints | 0 | 10, 70 | 20, 100 | 30, 120 |
|---|---|---|---|---|
| Finetune U-Net | | 107.16 | 143.02 | 174.15 |
| Finetune text encoder | 84.01 | 109.48 | 156.10 | 202.79 |
| Fusion | | 126.59 | 207.80 | 250.18 |

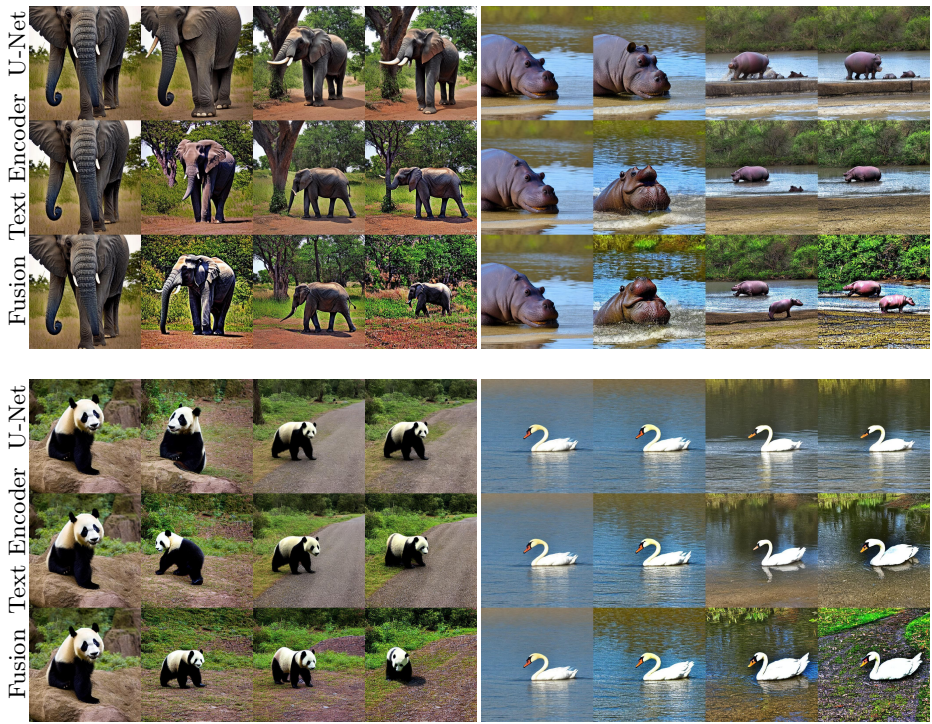

**Fig. 15:** Visual examples of combining U-Net and text encoder at different checkpoints in the incompression task.

### D.2   Results of Aesthetic Reward with Animal Prompts

*Aesthetic rewards.* Same as previous works, we also conduct experiments using the LAION Aesthetics Predictor [6] as the aesthetic reward function. It should be noted that the aesthetic reward does not consider the coherence with the prompt, and it is easy for the model to hack the reward as shown in [10].

We compare results with DDPO and AlignProb in Fig. 16. Both DDPO and our approach are trained with 10K samples, while AlignProb used early stop to avoid model collapse. Figure 16 presents results for three distinct animals: *jellyfish*, *penguin*, and *swan*. We can observe that both DDPO and AlignProb are over-optimized to the rewards, resulting in over stylized images to maximize aesthetic scores. For instance, DDPO tends to produce images characterized by a blurred yellow background, and AlignProb tends to generate images with over-saturated colors and unrealistic textures. In contrast, our method can generate more lifelike images that closely align with the provided prompts. For example, the composition of the jellyfish image is aesthetically pleasing, and the unrealistic features of the penguin and swan images have been rectified.

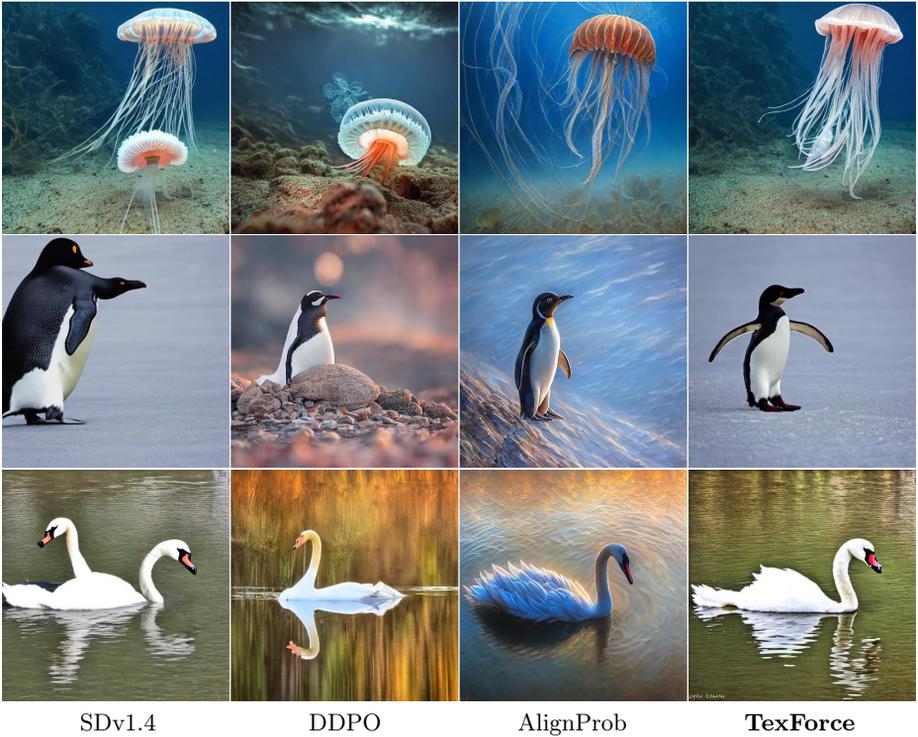

**Fig. 16:** Comparison with others on unseen animal prompts: *jellyfish*, *penguin* and *swan*.

**Table 6:** Quantitative results of ReFL-LoRA on the ImageReward dataset.

| Backbone | Original | ReFL | ReFL-LoRA | TexForce | ReFL+TexForce | ReFL-LoRA+TexForce |
|---|---|---|---|---|---|---|
| SDv1.4 | 0.2154 | 0.4485 | 0.4425 | 0.4556 | **0.6553** | **0.7093** |
| SDv1.5 | 0.2140 | 0.5484 | 0.5558 | 0.4086 | **0.6703** | **0.7438** |
| SDv2.1 | 0.3891 | 0.5223 | 0.5181 | 0.5084 | **0.6158** | **0.6263** |

### D.3 Training ReFL with LoRA

To make our model easier to use, we modified the original ReFL to train LoRA weights instead of the entire U-Net, and the results are shown in Tab. 6. We can notice that the performance of ReFL-LoRA is similar to ReFL but much better when combining with TexForce.

## E More Qualitative Results

In this section, we select more examples on different backbones in Figs. 17 to 22.

|SDv1.5 | ReFL | **TexForce** | ReFL+**TexForce**|

*Portrait of an old sea captain, male, detailed face, fantasy, highly detailed, cinematic, art painting by greg rutkowski*

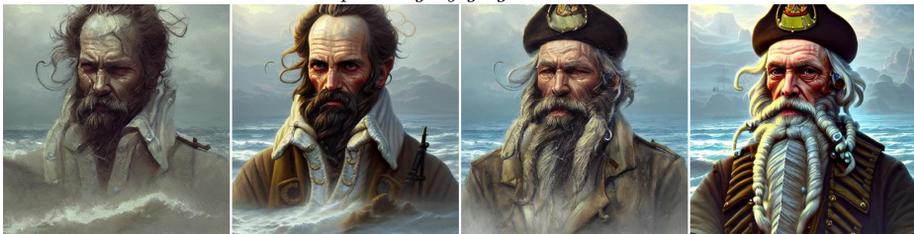

*A majestic landscape featuring a river, mountains and a forest, A small group of birds is flying in the sky, Harsh winter, very windy, There is a man walking in a deep snow, Cinematic, very, beautiful, painting, in the, style, of Lord of the rings*

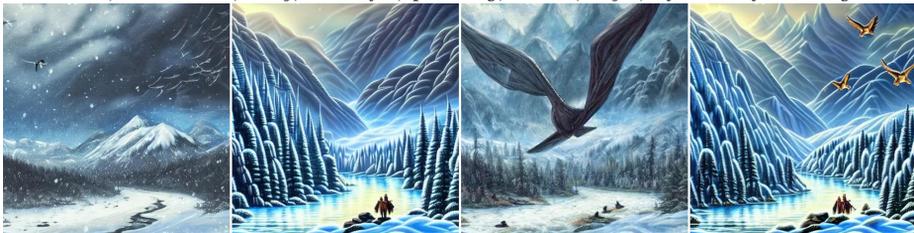

*astronaut drifting afloat in space, in the darkness away from anyone else, alone, black background dotted with stars, realistic*

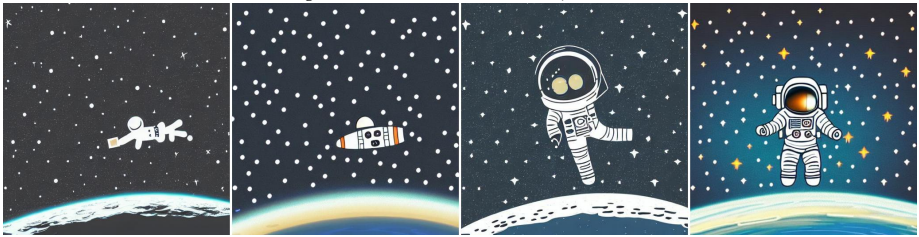

*plants, flowers, trees being mixed in a bowl*

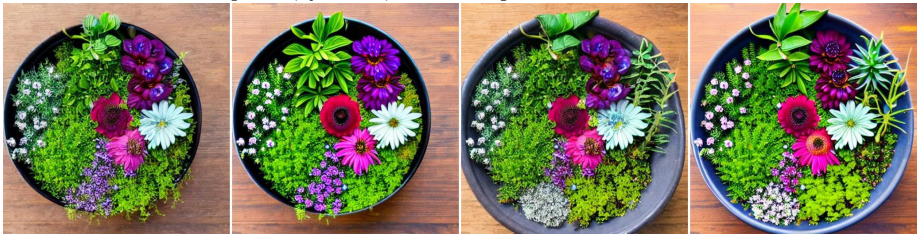

**Fig. 17:** Results of different methods with Stable Diffusion V1.4 as backbone on ImageReward test dataset.

## F   Limitations

Similar to other RL-based methods, our method also faces the challenges of sample efficiency and the complexity of reward function engineering.

| SDv1.5 | ReFL | **TexForce** | ReFL+**TexForce** |

*small red wooden cottage by the lake, lanterns on the porch, smoke coming out of the chimney, dusk, birch trees, tranquility, two swans swimming on the lake, a wooden rowing boat, cumulus clouds, by charlie bowater, by greg rutkowski*

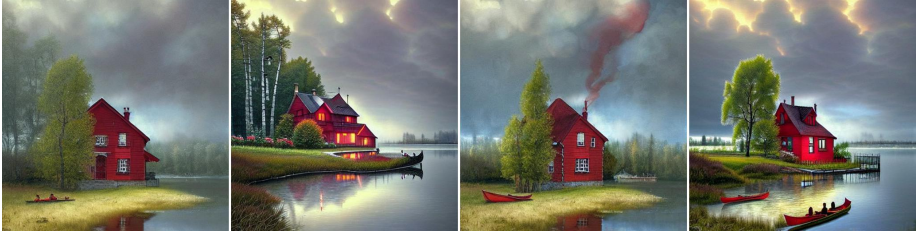

*portrait of a cute cyberpunk cat, realistic, professional*

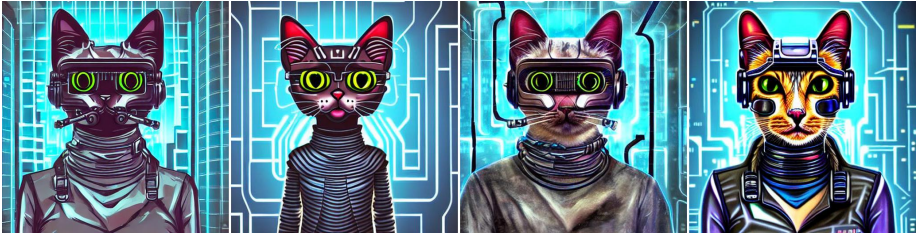

*field of light blue lotus flowers, minimalistic art, elegant*

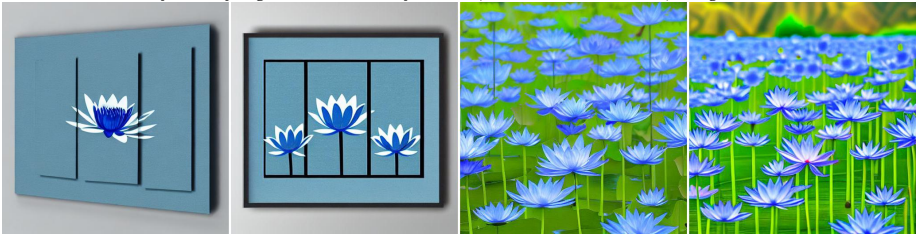

*a coffee mug made of cardboard*

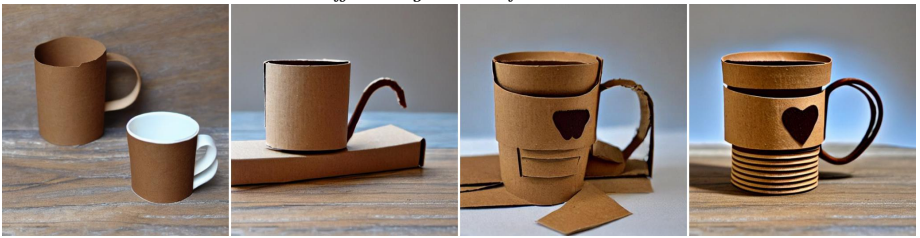

**Fig. 18:** Results of different methods with Stable Diffusion V1.5 as backbone on ImageReward test dataset.

## G  Broader Impacts

Since TexForce can finetune text-to-image models to satisfy specific rewards, it presents potential societal concerns regarding misinformation, intellectual property rights, and illegal usage of the model.

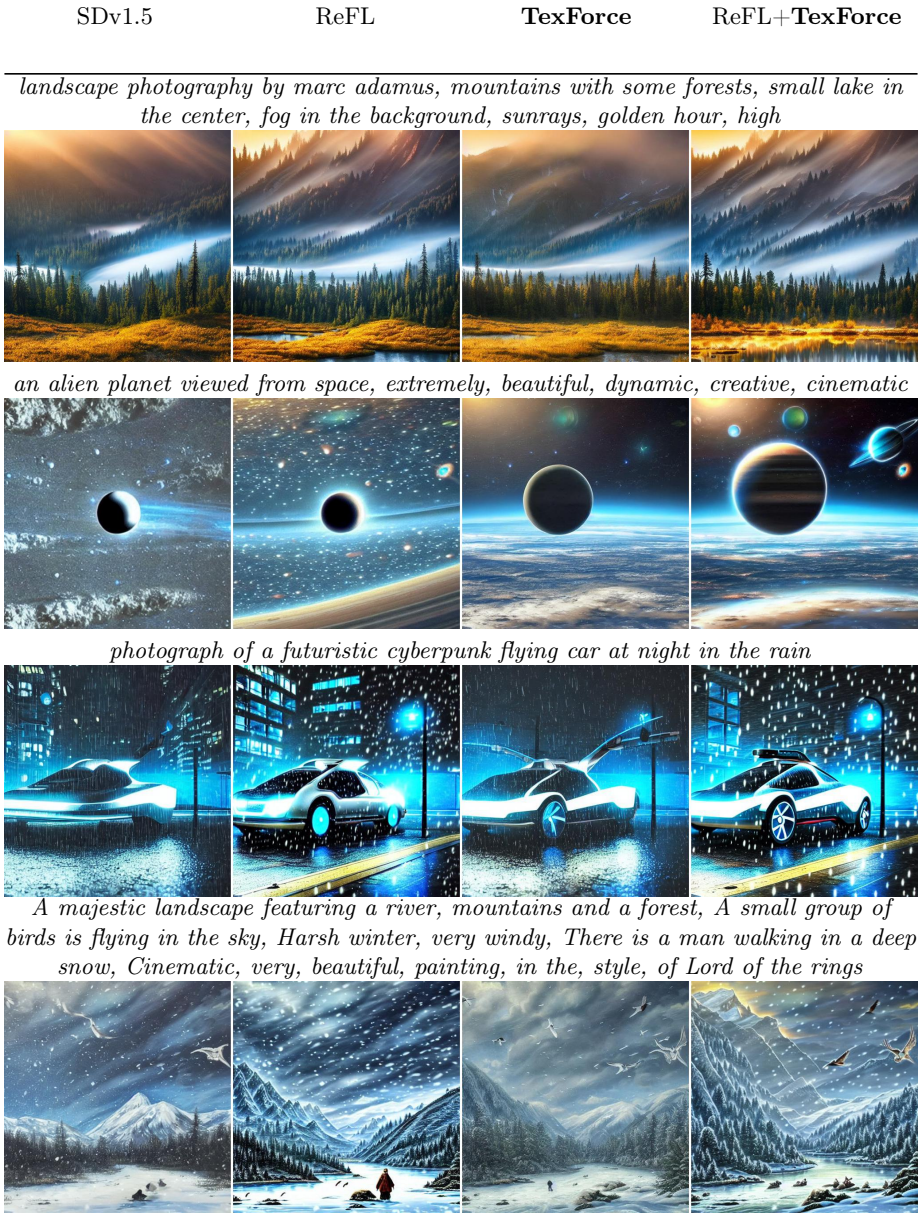

**Fig. 19:** Results of different methods with Stable Diffusion V1.5 as backbone on ImageReward test dataset.

|  SDv1.5 | ReFL | **TexForce** | ReFL+**TexForce** |

*extremely detailed stunning beautiful futuristic smooth curvilinear museum interior, colorful, hyper, real*

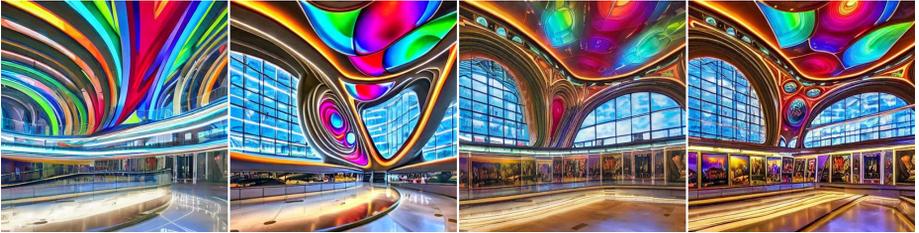

*beautiful prince, male, golden hair, high, fantasy, art by anato finnstark, joseph leyendecker, peter mohrbacher, ruan jia, marc simonetti, ayami kojima, cedric peyravernay, finnian macmanus, alphonse mucha, victo ngai*

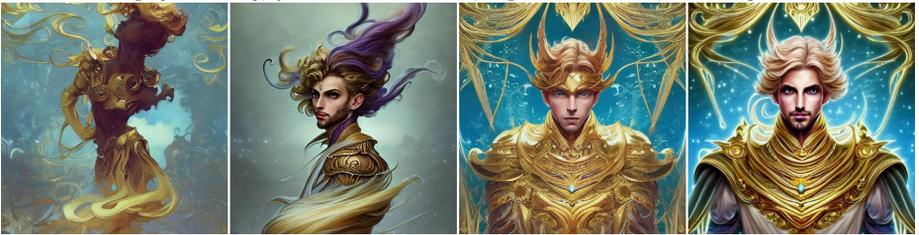

*small red wooden cottage by the lake, lanterns on the porch, smoke coming out of the chimney, dusk, birch trees, tranquility, two swans swimming on the lake, a wooden rowing boat, cumulus clouds, by charlie bowater, by greg rutkowski*

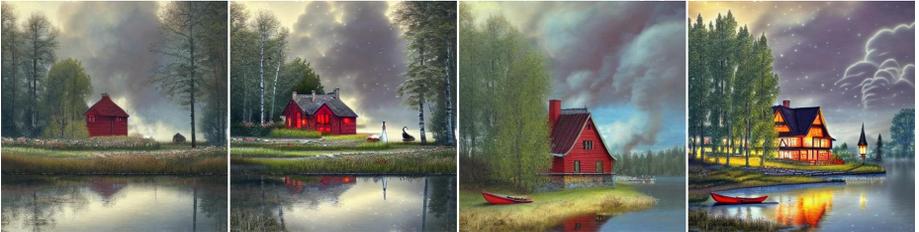

*natural suburb with multiple low rise apartment buildings in a park like setting with green hilly lawns and lush trees, pine wooden walls, rustic, large glass windows, cobblestone, grass, white, natural pathways, natural materials, minimalist, swedish design, bright, feng shui, modern, technology, frank lloyd wright*

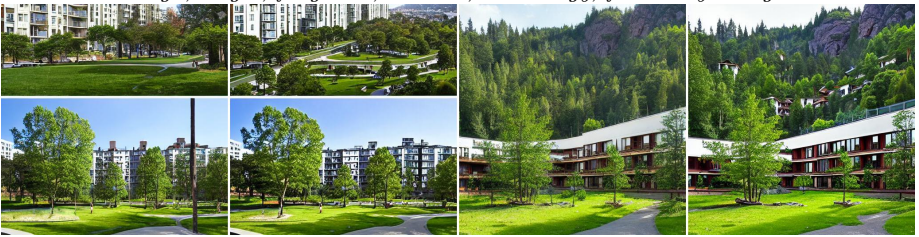

**Fig. 20:** Results of different methods with Stable Diffusion V1.5 as backbone on ImageReward test dataset.

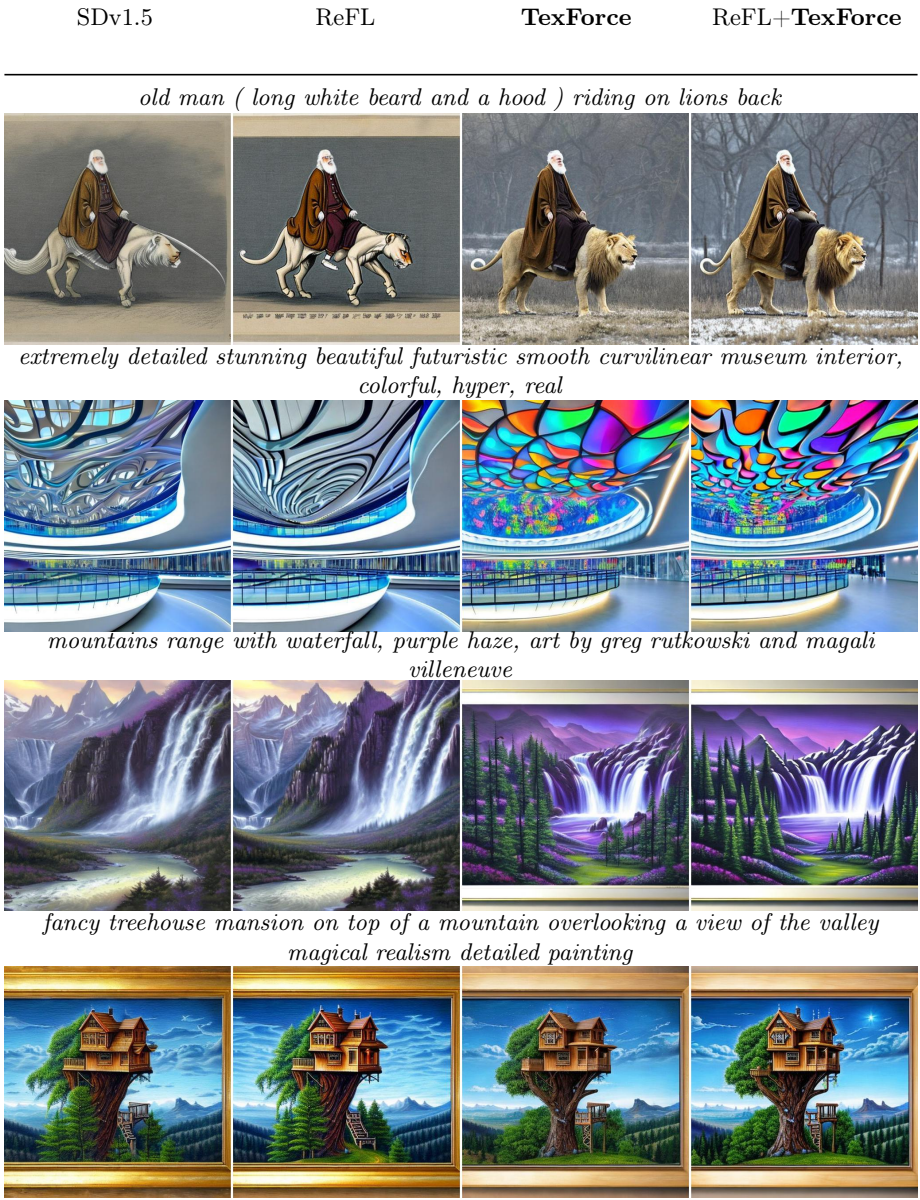

**Fig. 21:** Results of different methods with Stable Diffusion V2.1 as backbone on ImageReward test dataset.

| SDv1.5 | ReFL | **TexForce** | ReFL+**TexForce** |

*cinematic movie scene, beautiful Product shot film still of a Syd Mead futuristic modern sleek automobile speeding down a wet street at night in cyperpunk city, motion, hard surface modeling, soft, style of Stanley Kubrick cinematography*

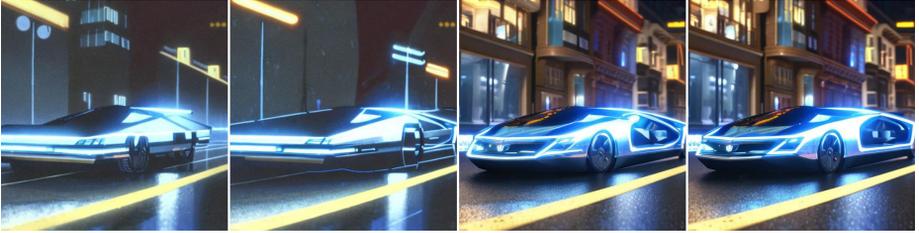

*field of light blue lotus flowers, minimalistic art, elegant*

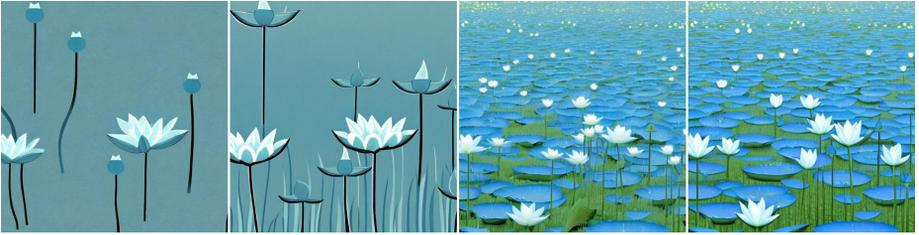

*the grand hall of the sacred library oil painting by james gurney*

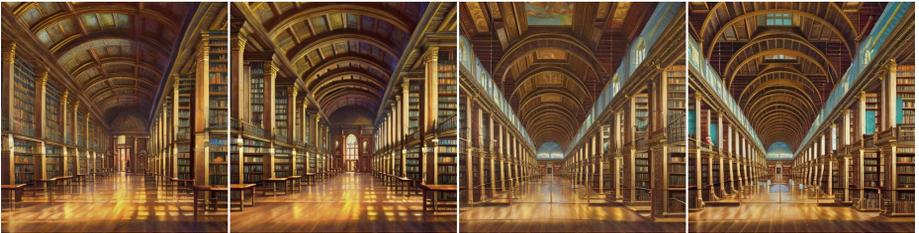

*hyperdetailed samsung store, oak parquet, black walls, digital walls, plants, light corona octane*

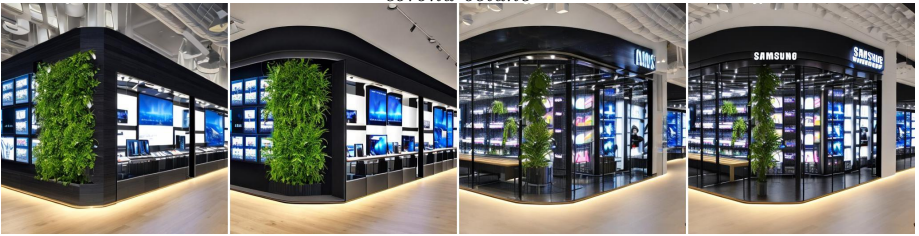

**Fig. 22:** Results of different methods with Stable Diffusion V2.1 as backbone on ImageReward test dataset.



\end{document}

%% file: custom_cmds.tex
\graphicspath{{./figures/}{./supp_figs/}}

\usepackage[group-separator={,}]{siunitx}
\usepackage{amsmath}
\usepackage{amsfonts}
\usepackage{makecell}
\usepackage{lipsum}
\usepackage{tcolorbox}
\usepackage{multirow}
\usepackage{wrapfig}
\usepackage{graphicx}
\usepackage{bbding}

\usepackage[capitalize]{cleveref}
\crefname{section}{Sec.}{Secs.}
\Crefname{section}{Section}{Sections}
\Crefname{table}{Table}{Tables}
\crefname{table}{Tab.}{Tabs.}
\Crefname{figure}{Figure}{Figures}
\crefname{figure}{Fig.}{Figs.}
\Crefname{equation}{Equation}{Equations}
\crefname{equation}{Eq.}{Eqs.}
\Crefname{algorithm}{Algorithm}{Algorithms}
\crefname{algorithm}{Alg.}{Algs.}

\usepackage[labelformat=simple]{subcaption}

\newcommand{\xo}{{\mathbf{x}_0}}
\newcommand{\mx}{{\mathbf{x}}}
\newcommand{\mz}{{\mathbf{z}}}

\newcommand{\nd}{{\mathcal{N}}}

\newcommand{\msmall}[1]{{\fontsize{8pt}{10pt}\selectfont #1}}

\newcommand{\para}[1]{{\noindent \textbf{#1}}}

\newcommand{\second}[1]{\textcolor{blue}{\textbf{#1}}}
\newcommand{\best}[1]{\textcolor{red}{\textbf{#1}}}

\colorlet{lightpink}{pink!35}
\colorlet{lightcyan}{cyan!20}
\colorlet{red}{red!80}
\colorlet{blue}{blue!80}
\colorlet{green}{green!60!black}
\colorlet{algemp}{cyan!10}

\usepackage{dirtree}
\usepackage{xcolor,colortbl}

\newcolumntype{a}{>{\columncolor{gray!20!white}}c}

%% file: main.bbl
\begin{thebibliography}{10}
\providecommand{\url}[1]{\texttt{#1}}
\providecommand{\urlprefix}{URL }
\providecommand{\doi}[1]{https://doi.org/#1}

\bibitem{Alam2021Unified}
Alam, M.M., Islam, M.T., Rahman, S.M.M.: Unified learning approach for
  egocentric hand gesture recognition and fingertip detection. Pattern
  Recognition  (2021). \doi{10.1016/j.patcog.2021.108200}

\bibitem{black2023ddpo}
Black, K., Janner, M., Du, Y., Kostrikov, I., Levine, S.: Training diffusion
  models with reinforcement learning. Int. Conf. Learn. Represent.  (2024)

\bibitem{brock2018largebiggan}
Brock, A., Donahue, J., Simonyan, K.: Large scale gan training for high
  fidelity natural image synthesis. Int. Conf. Learn. Represent.  (2019)

\bibitem{pyiqa}
Chen, C., Mo, J.: {IQA-PyTorch}: Pytorch toolbox for image quality assessment.
  [Online]. Available: \url{https://github.com/chaofengc/IQA-PyTorch} (2022)

\bibitem{chen2023topiq}
Chen, C., Mo, J., Hou, J., Wu, H., Liao, L., Sun, W., Yan, Q., Lin, W.: Topiq:
  A top-down approach from semantics to distortions for image quality
  assessment. In: IEEE Trans. Image Process. (2024)

\bibitem{laionaes}
Christoph, S., Romain, B.: Laion-aesthetics. (2022)

\bibitem{clark2023draft}
Clark, K., Vicol, P., Swersky, K., Fleet, D.J.: Directly fine-tuning diffusion
  models on differentiable rewards. Int. Conf. Learn. Represent.  (2024)

\bibitem{dai2023emu}
Dai, X., Hou, J., Ma, C.Y., Tsai, S., Wang, J., Wang, R., Zhang, P.,
  Vandenhende, S., Wang, X., Dubey, A., et~al.: Emu: Enhancing image generation
  models using photogenic needles in a haystack. arXiv preprint
  arXiv:2309.15807  (2023)

\bibitem{dhariwal2021diffusion}
Dhariwal, P., Nichol, A.: Diffusion models beat gans on image synthesis. In:
  Adv. Neural Inform. Process. Syst. (2021)

\bibitem{dong2023raft}
Dong, H., Xiong, W., Goyal, D., Pan, R., Diao, S., Zhang, J., Shum, K., Zhang,
  T.: Raft: Reward ranked finetuning for generative foundation model alignment.
  arXiv preprint arXiv:2304.06767  (2023)

\bibitem{2023DPOK}
Fan, Y., Watkins, O., Du, Y., Liu, H., Ryu, M., Boutilier, C., Abbeel, P.,
  Ghavamzadeh, M., Lee, K., Lee, K.: Dpok: Reinforcement learning for
  fine-tuning text-to-image diffusion models. Adv. Neural Inform. Process.
  Syst.  (2024)

\bibitem{gal2022textureinversion}
Gal, R., Alaluf, Y., Atzmon, Y., Patashnik, O., Bermano, A.H., Chechik, G.,
  Cohen-Or, D.: An image is worth one word: Personalizing text-to-image
  generation using textual inversion. Int. Conf. Learn. Represent.  (2023)

\bibitem{goodfellow2020generative}
Goodfellow, I., Pouget-Abadie, J., Mirza, M., Xu, B., Warde-Farley, D., Ozair,
  S., Courville, A., Bengio, Y.: Generative adversarial networks.
  Communications of the ACM  \textbf{63}(11),  139--144 (2020)

\bibitem{guo2023animatediff}
Guo, Y., Yang, C., Rao, A., Wang, Y., Qiao, Y., Lin, D., Dai, B.: Animatediff:
  Animate your personalized text-to-image diffusion models without specific
  tuning. Int. Conf. Learn. Represent.  (2024)

\bibitem{hao2022optimizing}
Hao, Y., Chi, Z., Dong, L., Wei, F.: Optimizing prompts for text-to-image
  generation. Adv. Neural Inform. Process. Syst.  (2023)

\bibitem{ho2020denoisingddpm}
Ho, J., Jain, A., Abbeel, P.: Denoising diffusion probabilistic models. In:
  Adv. Neural Inform. Process. Syst. (2020)

\bibitem{ho2022classifier}
Ho, J., Salimans, T.: Classifier-free diffusion guidance. arXiv preprint
  arXiv:2207.12598  (2022)

\bibitem{lora}
Hu, E.J., Shen, Y., Wallis, P., Allen-Zhu, Z., Li, Y., Wang, S., Wang, L.,
  Chen, W.: Lora: Low-rank adaptation of large language models. Int. Conf.
  Learn. Represent.  (2022)

\bibitem{tifa}
Hu, Y., Liu, B., Kasai, J., Wang, Y., Ostendorf, M., Krishna, R., Smith, N.A.:
  Tifa: Accurate and interpretable text-to-image faithfulness evaluation with
  question answering. Int. Conf. Comput. Vis.  (2023)

\bibitem{karras2019style}
Karras, T., Laine, S., Aila, T.: A style-based generator architecture for
  generative adversarial networks. In: IEEE Conf. Comput. Vis. Pattern Recog.
  pp. 4401--4410 (2019)

\bibitem{text2video-zero}
Khachatryan, L., Movsisyan, A., Tadevosyan, V., Henschel, R., Wang, Z.,
  Navasardyan, S., Shi, H.: Text2video-zero: Text-to-image diffusion models are
  zero-shot video generators. Int. Conf. Comput. Vis.  (2023)

\bibitem{pickscore}
Kirstain, Y., Polyak, A., Singer, U., Matiana, S., Penna, J., Levy, O.:
  Pick-a-pic: An open dataset of user preferences for text-to-image generation.
  Adv. Neural Inform. Process. Syst.  (2024)

\bibitem{lee2023aligning}
Lee, K., Liu, H., Ryu, M., Watkins, O., Du, Y., Boutilier, C., Abbeel, P.,
  Ghavamzadeh, M., Gu, S.S.: Aligning text-to-image models using human
  feedback. arXiv preprint arXiv:2302.12192  (2023)

\bibitem{agiqa3k}
Li, C., Zhang, Z., Wu, H., Sun, W., Min, X., Liu, X., Zhai, G., Lin, W.:
  Agiqa-3k: An open database for ai-generated image quality assessment. IEEE
  Trans. Circuit Syst. Video Technol. pp.~1--1 (2023).
  \doi{10.1109/TCSVT.2023.3319020}

\bibitem{li2023w-plus-adapter}
Li, X., Hou, X., Loy, C.C.: When stylegan meets stable diffusion: a
  $\mathcal{W}_+$ adapter for personalized image generation. IEEE Conf. Comput.
  Vis. Pattern Recog.  (2024)

\bibitem{li2024textcraftor}
Li, Y., Liu, X., Kag, A., Hu, J., Idelbayev, Y., Sagar, D., Wang, Y., Tulyakov,
  S., Ren, J.: Textcraftor: Your text encoder can be image quality controller.
  In: IEEE Conf. Comput. Vis. Pattern Recog. pp. 7985--7995 (2024)

\bibitem{liu2022character}
Liu, R., Garrette, D., Saharia, C., Chan, W., Roberts, A., Narang, S., Blok,
  I., Mical, R., Norouzi, M., Constant, N.: Character-aware models improve
  visual text rendering. arXiv preprint arXiv:2212.10562  (2022)

\bibitem{liu2023zero}
Liu, R., Wu, R., Van~Hoorick, B., Tokmakov, P., Zakharov, S., Vondrick, C.:
  Zero-1-to-3: Zero-shot one image to 3d object. In: Int. Conf. Comput. Vis.
  pp. 9298--9309 (2023)

\bibitem{mou2023t2i}
Mou, C., Wang, X., Xie, L., Wu, Y., Zhang, J., Qi, Z., Shan, Y., Qie, X.:
  T2i-adapter: Learning adapters to dig out more controllable ability for
  text-to-image diffusion models. AAAI  (2024)

\bibitem{ouyang2022training}
Ouyang, L., Wu, J., Jiang, X., Almeida, D., Wainwright, C., Mishkin, P., Zhang,
  C., Agarwal, S., Slama, K., Ray, A., et~al.: Training language models to
  follow instructions with human feedback. In: Adv. Neural Inform. Process.
  Syst. (2022)

\bibitem{poole2022dreamfusion}
Poole, B., Jain, A., Barron, J.T., Mildenhall, B.: Dreamfusion: Text-to-3d
  using 2d diffusion. Int. Conf. Learn. Represent.  (2023)

\bibitem{prabhudesai2023alignprop}
Prabhudesai, M., Goyal, A., Pathak, D., Fragkiadaki, K.: Aligning text-to-image
  diffusion models with reward backpropagation. arXiv preprint arXiv:2310.03739
   (2023)

\bibitem{clip}
Radford, A., Kim, J.W., Hallacy, C., Ramesh, A., Goh, G., Agarwal, S., Sastry,
  G., Askell, A., Mishkin, P., Clark, J., et~al.: Learning transferable visual
  models from natural language supervision. In: Proc. Int. Conf. Mach. Learn.
  pp. 8748--8763 (2021)

\bibitem{dalle2}
Ramesh, A., Dhariwal, P., Nichol, A., Chu, C., Chen, M.: Hierarchical
  text-conditional image generation with clip latents. arXiv preprint
  arXiv:2204.06125  (2022)

\bibitem{dalle1}
Ramesh, A., Pavlov, M., Goh, G., Gray, S., Voss, C., Radford, A., Chen, M.,
  Sutskever, I.: Zero-shot text-to-image generation. In: Proc. Int. Conf. Mach.
  Learn. (2021)

\bibitem{stablediffusion}
Rombach, R., Blattmann, A., Lorenz, D., Esser, P., Ommer, B.: High-resolution
  image synthesis with latent diffusion models. In: IEEE Conf. Comput. Vis.
  Pattern Recog. (2022)

\bibitem{ruiz2023dreambooth}
Ruiz, N., Li, Y., Jampani, V., Pritch, Y., Rubinstein, M., Aberman, K.:
  Dreambooth: Fine tuning text-to-image diffusion models for subject-driven
  generation. In: IEEE Conf. Comput. Vis. Pattern Recog. pp. 22500--22510
  (2023)

\bibitem{imagen}
Saharia, C., Chan, W., Saxena, S., Li, L., Whang, J., Denton, E., Ghasemipour,
  S.K.S., Ayan, B.K., Mahdavi, S.S., Lopes, R.G., et~al.: Photorealistic
  text-to-image diffusion models with deep language understanding. In: Adv.
  Neural Inform. Process. Syst. (2022)

\bibitem{laion-5b}
Schuhmann, C., Beaumont, R., Vencu, R., Gordon, C., Wightman, R., Cherti, M.,
  Coombes, T., Katta, A., Mullis, C., Wortsman, M., et~al.: Laion-5b: An open
  large-scale dataset for training next generation image-text models. arXiv
  preprint arXiv:2210.08402  (2022)

\bibitem{laion400m}
Schuhmann, C., Vencu, R., Beaumont, R., Kaczmarczyk, R., Mullis, C., Katta, A.,
  Coombes, T., Jitsev, J., Komatsuzaki, A.: Laion-400m: Open dataset of
  clip-filtered 400 million image-text pairs. arXiv preprint arXiv:2111.02114
  (2021)

\bibitem{ppo}
Schulman, J., Wolski, F., Dhariwal, P., Radford, A., Klimov, O.: Proximal
  policy optimization algorithms. arXiv preprint arXiv:1707.06347  (2017)

\bibitem{sohl2015deep}
Sohl-Dickstein, J., Weiss, E., Maheswaranathan, N., Ganguli, S.: Deep
  unsupervised learning using nonequilibrium thermodynamics. In: Proc. Int.
  Conf. Mach. Learn. (2015)

\bibitem{ddim}
Song, J., Meng, C., Ermon, S.: Denoising diffusion implicit models. Int. Conf.
  Learn. Represent.  (2021)

\bibitem{song2021maximum}
Song, Y., Durkan, C., Murray, I., Ermon, S.: Maximum likelihood training of
  score-based diffusion models. Adv. Neural Inform. Process. Syst.
  \textbf{34},  1415--1428 (2021)

\bibitem{song2020improved}
Song, Y., Ermon, S.: Improved techniques for training score-based generative
  models. In: Adv. Neural Inform. Process. Syst. (2020)

\bibitem{gfiqa20k}
Su, S., Lin, H., Hosu, V., Wiedemann, O., Sun, J., Zhu, Y., Liu, H., Zhang, Y.,
  Saupe, D.: Going the extra mile in face image quality assessment: A novel
  database and model. IEEE Trans. Multimedia  (2023)

\bibitem{wang2022diffusiondb}
Wang, Z.J., Montoya, E., Munechika, D., Yang, H., Hoover, B., Chau, D.H.:
  Diffusiondb: A large-scale prompt gallery dataset for text-to-image
  generative models. arXiv preprint arXiv:2210.14896  (2022)

\bibitem{witteveen2022investigating}
Witteveen, S., Andrews, M.: Investigating prompt engineering in diffusion
  models. arXiv preprint arXiv:2211.15462  (2022)

\bibitem{wu2023qbench}
Wu, H., Zhang, Z., Zhang, E., Chen, C., Liao, L., Wang, A., Li, C., Sun, W.,
  Yan, Q., Zhai, G., et~al.: Q-bench: A benchmark for general-purpose
  foundation models on low-level vision. Int. Conf. Learn. Represent.  (2024)

\bibitem{wu2023qinstruct}
Wu, H., Zhang, Z., Zhang, E., Chen, C., Liao, L., Wang, A., Xu, K., Li, C.,
  Hou, J., Zhai, G., Xue, G., Sun, W., Yan, Q., Lin, W.: Q-instruct: Improving
  low-level visual abilities for multi-modality foundation models. IEEE Conf.
  Comput. Vis. Pattern Recog.  (2024)

\bibitem{wu2023tune}
Wu, J.Z., Ge, Y., Wang, X., Lei, S.W., Gu, Y., Shi, Y., Hsu, W., Shan, Y., Qie,
  X., Shou, M.Z.: Tune-a-video: One-shot tuning of image diffusion models for
  text-to-video generation. In: Int. Conf. Comput. Vis. pp. 7623--7633 (2023)

\bibitem{hps}
Wu, X., Sun, K., Zhu, F., Zhao, R., Li, H.: Better aligning text-to-image
  models with human preference. Int. Conf. Comput. Vis.  (2023)

\bibitem{imagereward}
Xu, J., Liu, X., Wu, Y., Tong, Y., Li, Q., Ding, M., Tang, J., Dong, Y.:
  Imagereward: Learning and evaluating human preferences for text-to-image
  generation. Adv. Neural Inform. Process. Syst.  (2023)

\bibitem{zhang2023controlnet}
Zhang, L., Rao, A., Agrawala, M.: Adding conditional control to text-to-image
  diffusion models. In: Int. Conf. Comput. Vis. (2023)

\end{thebibliography}
